\documentclass[lettersize,journal]{IEEEtran}
\usepackage{cite}
\usepackage{amsmath,amssymb,amsfonts}
\usepackage{algorithmic}
\usepackage{algorithm}
\usepackage{flushend}
\usepackage{array}
\usepackage{cuted}
\usepackage[caption=false,font=normalsize,labelfont=sf,textfont=sf]{subfig}
\usepackage{textcomp}
\usepackage{stfloats}
\usepackage{url}
\usepackage{verbatim}
\usepackage{graphicx}
\usepackage{hyperref}
\usepackage{multirow}
\usepackage{booktabs}
\usepackage{color}
\usepackage{enumerate}
\begin{document}

\title{Self-supervised remote sensing feature learning: Learning Paradigms, Challenges, and Future Works}

\author{Chao Tao \IEEEmembership{Member, IEEE}, Ji Qi, Mingning Guo, Qing Zhu and Haifeng Li* \IEEEmembership{Member, IEEE}
\thanks{Manuscript received XXX, 2022; revised XXX, 2022. (Corresponding author: H.F. Li)}
\thanks{C. Tao, J. Qi, M.N. Guo and H.F. Li are with the School of Geosciences and Info-Physics, Central South University, Changsha 410083, China (e-mail: kingtaochao@csu.edu.cn; jameschi95@foxmail.com; lihaifeng@csu.edu.cn).}
\thanks{Q. Zhu is with the Faculty of Geosciences and Environmental Engineering, Southwest Jiaotong University, Chengdu 611756, China (e-mail: zhuq66@263.net).}
}

\markboth{Journal of \LaTeX\ Class Files,~Vol.~14, No.~8, August~2021}%
{Shell \MakeLowercase{\textit{et al.}}: A Sample Article Using IEEEtran.cls for IEEE Journals}

\IEEEpubid{0000--0000/00\$00.00~\copyright~2021 IEEE}

\maketitle

\begin{abstract}
Deep learning has achieved great success in learning features from massive remote sensing images (RSIs). To better understand the connection between feature learning paradigms (e.g., unsupervised feature learning (USFL), supervised feature learning (SFL), and self-supervised feature learning (SSFL)), this paper analyzes and compares them from the perspective of feature learning signals, and gives a unified feature learning framework. Under this unified framework, we analyze the advantages of SSFL over the other two learning paradigms in RSIs understanding tasks and give a comprehensive review of the existing SSFL work in RS, including the pre-training dataset, self-supervised feature learning signals, and the evaluation methods. We further analyze the effect of SSFL signals and pre-training data on the learned features to provide insights for improving the RSI feature learning. Finally, we briefly discuss some open problems and possible research directions.
\end{abstract}


\section{Introduction}\label{sec:introduction}

\IEEEPARstart{O}{btaining} a feature representation being both discriminative and invariable is fundamental in remote sensing image (RSI) understanding \cite{Zhang_Zhang_Du_2016, Zhu_Tuia_Mou_Xia_Zhang_Xu_Fraundorfer_2017, Yuan_Shen_Li_Li_Li_Jiang_Xu_Tan_Yang_Wang_2020}. The feature discrimination concern is closely related to the visual manifold hypothesis \cite{Fefferman_Mitter_Narayanan_2016}, that is, the real-world data samples presented in high dimensional spaces are usually embedded into a lower-dimensional manifold in human vision perception, where samples from different categories are naturally clustered. In other words, visual separability is highly dependent on the discriminative power of feature representation. The advantage of human vision lies in the invariance of object recognition under various environments. In the field of RSI understanding, the visual feature of objects is generally influenced by both external and internal factors \cite{Cui_Zhang_Wang_Li_Qi_2021, Wang_Tao_Qi_Xiao_Li_2022}. External factors are the variation of imaging condition (e.g., illumination, viewpoint, and scale) and imaging mechanism (e.g., optical and SAR image), and the internal factors are the internal change over time like vegetation phenological shifts. These factors lead to significant changes in visual features, which makes remote sensing image understanding tasks very challenging. Thus, obtaining the feature representation insensitive to changes has received much attention, which is also the key to ensuring the generalization ability of the RSI understanding model for different regions, different times, and different imaging conditions.

Instead of traditional hand-crafted feature engineering, learning feature representation from massive data has attracted great attention during the past decade, and exhibited very impressive performance in many RSI understanding tasks \cite{Xia_Hu_Hu_Shi_Bai_Zhong_Zhang_Lu_2017, Cheng_Han_Lu_2017, Xia_Bai_Ding_Zhu_Belongie_Luo_Datcu_Pelillo_Zhang_2018, Lu_Tao_Li_Qi_Li_2022, Tao_Lu_Qi_Wang_2021, Chen_Li_Bai_Yang_Jiang_Miao_2021}. The success of feature learning is mainly attributed to two factors. First, it uses deep hierarchical network architectures. In 2012, Bengio et al. \cite{Bengio_Courville_Vincent_2013} discussed that the key to a good representation is to use deep architectures for feature learning, which carries two advantages: 1) deep architectures reuse features; 2) deep architectures help to learn abstract semantic features that keep invariant under most changes of the input. Second, it uses a large number of samples for deep network model training. Sun et al. \cite{Sun_Shrivastava_Singh_Gupta_2017} performed an empirical study using the JFT-300M dataset containing 300M samples for representation learning. They found that the performance of feature representation increases linearly with the number of training data for high-capacity models. In the remote sensing community, Long et al. \cite{Long_Xia_Li_Yang_Yang_Zhu_Zhang_Li_2021} compared the performance of feature learning trained on existing remote sensing benchmark datasets. They argued that using a small number of samples may limit the performance of feature learning because a large number of remote sensing samples is more likely to represent the feature distribution of land cover in the real world.

Early studies on feature learning aimed at learning features in an unsupervised way using models like sparse coding \cite{Cheriyadat_2014, Dai_Yang_2011, Ratha_Bhattacharya_Frery_2018, Romero_Gatta_Camps-Valls_2016, Sheng_Yang_Xu_Sun_2012}, autoencoder \cite{Vincent_Larochelle_Lajoie_Bengio_Manzagol_2010, Chao_Pan_Li_Zou_2015, Tao_Xu_Zhang_Du_Zhang_2017, Kemker_Kanan_2017, Mou_Ghamisi_Zhu_2018, Ozkan_Kaya_Akar_2019, Tomenotti_Luppino_Hansen_Moser_Anfinsen_2020, Jin_Ma_Fan_Huang_Mei_Ma_2021, Sharma_Hara_2021, Palsson_Sveinsson_Ulfarsson_2022}, and deep belief networks \cite{Lu_Zheng_Yuan_2017}. However, due to the lack of ground truth and an effective feedback mechanism in the learning process, the learned features may not be both discriminative and invariable enough for RSI understanding tasks. With the development of convolutional neural networks (CNNs), many supervised feature learning methods were proposed and have proved to be promising in extracting high-level visual features with distinguishability and invariance, and are successfully applied to RSI understanding tasks. Though supervised feature learning (SFL) paradigms have witnessed great progress over unsupervised feature learning (USFL) paradigms, they require large-scale, high-quality labeled data, which are difficult to obtain as accurately annotating RSIs is tedious and requires rich experience and geographic knowledge.
Moreover, the annotating approach used for RSIU tasks is extremely task-dependent. For example, scene classification tasks require image-level annotation, while semantic segmentation tasks require pixel-level annotation. In other words, different tasks need different level annotations, \\ \hspace*{\fill} \\ so great efforts need to be paid to constructing datasets in a task-dependent way.


Unlike machine vision which is “taught” by labeled data, human-like vision is not limited to a specific task or specific dataset, and human language-based labels are not the prerequisite for constructing the human visual system. Thus, a new feature learning paradigm, self-supervised feature learning (SSFL), was proposed in the field of natural language process (NLP) \cite{Radford_Narasimhan_Salimans_Sutskever_2018, Devlin_Chang_Lee_Toutanova_2019, Radford_Wu_Child_Luan_Amodei_Sutskever_2019, Brown_Mann_Ryder_Subbiah_Kaplan_Dhariwal_Neelakantan_Shyam_Sastry_Askell_2020} and computer vision (CV) \cite{Kolesnikov_Zhai_Beyer_2019, Newell_Deng_2020, Radford_Kim_Hallacy_Ramesh_Goh_Agarwal_Sastry_Askell_Mishkin_Clark_2021}. It uses human-designed task-agnostic self-supervised learning signals to generate pseudo-labels for massive unlabeled data, thereby replacing human labels to guide the model learning.

Up to now, several surveys on SSFL in computer vision have been done \cite{Liu_Zhang_Hou_Mian_Wang_Zhang_Tang_2021, Jing_Tian_2021, Ericsson_Gouk_Loy_Hospedales_2022, Wang_Albrecht_Braham_Mou_Zhu_2022}. To bridge the gap between the progress of SSFL in computer vision and remote sensing, Wang et al. \cite{Wang_Albrecht_Braham_Mou_Zhu_2022} summarized some representative methods of SSFL and analyzed their application in remote sensing tasks. Here, we want to further contribute to the study on SSFL for remote sensing applications as follows:

\begin{enumerate}
\item{For learning feature signals, we introduce a unified feature learning framework to understand the connection and difference between the SSFL paradigms and other feature learning paradigms.
}
\item{Based on this unified framework, we propose a new taxonomy of SSFL in terms of data, self-supervised feature learning signals, and evaluation methods, and perform an extensive literature review of the remote sensing community in these three aspects. Moreover, we comprehensively compare the properties of training data and self-supervised learning signals to provide a deeper understanding of the necessary conditions for better self-supervised learning for RSI.
}
\item{We discuss the limitations of existing remote sensing SSFL and suggest potential research directions including new ones that are not covered by previous surveys.
}
\end{enumerate}

The rest of this survey is organized as follows. Section \ref{sec:paradigm} introduces the mainstream feature learning paradigms in RS, and clarify the advantages of SSFL over other feature learning paradigms. Section \ref{sec:ssl} reviews the existing studies on SSFL in remote sensing, considering the training data, SSFL signals, and feature evaluation methods. Section \ref{sec:exp} analyzes how the SSFL signal and properties of pre-training data affect the performance of the learned features in downstream tasks. In Section \ref{sec:future_work}, we suggest four potential research directions of SSFL in remote sensing. Some conclusions are drawn in Section \ref{sec:conclusion}.

\section{A Unified Feature Learning Paradigm for Feature Learning Signals}\label{sec:paradigm}
\subsection{Definition}\label{subsec:definition}
Instead of handcrafted feature engineering, learning feature representation from massive data has been the mainstream feature extraction method due to its impressive performance in many RSI understanding tasks. In terms of the learning paradigm, feature learning methods can be divided into three categories: unsupervised feature learning (USFL), supervised feature learning (SFL), and self-supervised feature learning (SSFL). To understand the intrinsic relationship between these learning paradigms, we analyze these paradigms from the perspective of feature learning signals and proposed a unified feature learning framework.
Feature learning paradigm includes four parts: data, model, loss function, and optimizer, which can be mathematically described as:
\begin{equation}
\mathop{min}_{\theta}\mathcal{L}\left(f_\theta(\cdot), D\right).
\end{equation}
where $f_\theta(\cdot)$ is the model used for feature learning and $\theta $ is the parameters needed to be learned in the model. $f_\theta(\cdot)$ can be a deep convolutional model, auto-encoder model, Gaussian probability distribution model, etc.; $D$ denotes the samples used to train the model, which can be labeled or unlabeled datasets; $\mathcal{L}$ is the loss function, such as L2 loss and cross-entropy loss, which describes the metric used to approximate the ground truth or predefined optimization objective; $\mathop{min}_{\theta}$ denotes the optimizer, such as stochastic gradient descent (SGD) and evolutionary algorithm (EA), used to find the optimal parameters of the model.

In the SFL paradigm, the training sample set $D$ is usually denoted as $D={\left\{\left(x_i,y_i\right)|x_i\in {X},y_i\in {Y}\right\}}^N_{i=0}$, where $x_i$ and $y_i$ are the $i$th data sample and its corresponding label. $y_i$ is regarded as the ground truth of $x_i$ to guide the feature learning process, but this is set by humans according to their criteria. That is, if we find a learning signal hidden in the data as the ground truth, we can perform feature learning without labeled samples. We call this learning signal the generic supervised learning signal, and define a unified feature learning framework as follows:
\begin{equation}
\mathop{min}_{\theta}\mathcal{L}\left(f_\theta(\cdot), D, S\right).
\end{equation}
where $S$ denotes the generic supervised learning signal. In the following, we will explain how to define existing feature learning paradigms under this unified feature learning framework.

\subsection{Unsupervised feature learning}
From the perspective of learning signals, USFL uses the data intrinsic structure as the learning signal to guide the feature learning process. Specifically, given an unlabeled dataset $D={\{x_i|x_i\in {X}\}}^N_{i=0}$, for each $x_i$, a pseudo label $\widetilde{y_i}$ is constructed using the data intrinsic structure as a generic supervised learning signal. Thus, USFL can be defined as follows:
\begin{equation}
\mathop{min}_{\theta}\mathcal{L}(f_\theta(\cdot),\{(x_i,{\widetilde{Y}}_i)|x_i\in X,{\widetilde{Y}}_i\in \widetilde{Y}{\}}^N_{i=0},S).
\end{equation}
where $S: X \rightarrow \widetilde{Y}$. The commonly used USFL methods for remote sensing include the sparse coding model and autoencoder model. Specifically, the sparse coding model uses a complete visual dictionary learned from a large amount of unlabeled data to guide the reconstruction of the original image to obtain the corresponding sparse representation. The autoencoder is an encoder-decoder network that seeks to learn a compressed sparse representation of input by minimizing the reconstruction error between the input and output data. Although the two models use different algorithms, they both are based on the distribution law of data manifolds, that is, high-dimensional data of the same category tend to be concentrated near a low-dimensional manifold. However, due to the lack of ground truth or an effective feedback mechanism related to the specific RSI understanding task in the learning process, the learned features may not be both discriminative and invariable enough for RSI understanding tasks.

\subsection{Supervised feature learning}
In the SFL paradigm, the training samples $D$ are $D={\left\{(x_i, y_i)|x_i\in {X}, y_i\in {Y}\right\}}^N_{i=0}$, where $x_i$ denoting the $i$th data sample and $y_i$ denoting the corresponding data label. The data label $y_i$ can be regarded as the ground truth, which is set by human knowledge. Thus, the learning signal in the SFL paradigm is considered to be generated by human knowledge and SFL can be defined as follows:
\begin{equation}
\mathop{min}_{\theta}\mathcal{L}(f_\theta(\cdot),\{(x_i,y_i)|x_i\in X,y_i\in Y{\}}^N_{i=0},S).
\end{equation}
where $S: X \rightarrow Y$. The deep convolutional neural network (DCNN) is a typical method of SFL, which has achieved great success in RSI understanding tasks due to its advantages in representation learning. However, this method requires a large number of high-quality labeled data, which are extremely expensive to acquire, due to the spatial-temporal heterogeneity of remote sensing data. This contradiction seriously restricts the application of DCNN in large-scale and complex remote sensing image understanding tasks and brings two problems as follows:

\begin{enumerate}
\item From the perspective of training data, the successful application of DCNNs in RSI understanding highly depends on strong supervision, i.e., a large quantity of labeled training data. However, building a large high-quality training dataset is challenging: First, the representations of ground objects in RSIs vary with weather, climate, lighting, season, and satellite imaging condition. That is, the representation is affected by the temporal heterogeneity of RSIs. Each training sample only represents the characteristics of the objects at a specific time point, so building the RSI dataset for learning generic representations requires almost infinite labeled samples. Second, the distribution of ground objects varies over regions due to different climates and human activities \cite{Gong_Wang_Yu_Zhao_Zhao_Liang_Niu_Huang_Fu_Liu_2013, Gong_Liu_Zhang_Li_Wang_Huang_Clinton_Ji_Li_Bai_2019, Liu_Gong_Wang_Clinton_Bai_Liang_2020, Wang_Tao_Qi_Xiao_Li_2022}. Such spatial heterogeneity leads to sample category imbalance inside the training set or between the training and test sets during supervised learning, which leads to "over-learning" or "under-learning" problems in application \cite{Lopez_Fernandez_Garcia_Palade_Herrera_2013, Tuia_Persello_Bruzzone_2016}.

\item From the perspective of the learning mechanism, existing supervised learning trains the model on a limited number of samples, resulting in the contradiction between the closed sample set and the dynamics and complexity of ground object features, which causes the performance collapse of the model \cite{Shen_Liu_He_Zhang_Xu_Yu_Cui_2021, Zhang_Gao_2022, Zhou_Liu_Qiao_Xiang_Loy_2021}. Although this problem can be alleviated by increasing the number of labeled samples, the extremely high cost of obtaining high-quality data labels makes it difficult to model temporal heterogeneity. This is the inherent flaw of the supervised learning paradigm. Furthermore, supervised learning takes the semantic support provided by labels as the only learning signal. If labels are treated as a kind of prior knowledge, the model will be restricted in the given knowledge during the learning process. However, the large amount of remote sensing data contains far more information than that provided by sparse labels, theoretically. Therefore, over-reliance on manual labeling may lead to an "inductive bias" problem.
\end{enumerate}

\subsection{Self-supervised feature learning}\label{sec:paradigm_ssl}
SSFL uses human-designed task-agnostic learning signals to generate pseudo-labels for massive unlabeled data, thereby replacing human labels to guide the model learning. Specifically, given an unlabeled dataset $D={\{x_i | x_i\in X\}}^N_{i=0}$, for each $x_i$ in ${X}$, a pseudo label $\widetilde{y_i}$ is generated by human-designed learning signals. Thus, SSFL can also be defined by the unified feature learning framework as follows:
\begin{equation}
\mathop{min}_{\theta}\mathcal{L}(f_\theta(\cdot), \{(x_i, \widetilde{y_i})|x_i \in X, \widetilde{y_i} \in \widetilde{Y} \}^N_{i=0}, S).
\end{equation}
where $S: X \rightarrow \widetilde{Y}$. Although the USFL method and SSFL both use pseudo-labels for feature learning, the former constructs pseudo-labels using only the manifold hypothesis, but the latter use diverse pretext tasks (e.g., image inpainting, colorization, jigsaw). Thus, many studies have shown that the feature learned in a self-supervised manner is more powerful than the one learned in an unsupervised manner \cite{Schmarje_Santarossa_Schroder_Koch_2021}.

Recently, SSFL has got significant advances in natural language processing (NLP) \cite{Kayser_Camburu_Salewski_Emde_Do_Akata_Lukasiewicz_2021, Wang_Singh_Michael_Hill_Levy_Bowman_2018, Wang_Pruksachatkun_Nangia_Singh_Michael_Hill_Levy_Bowman_2019} and computer vision (CV) \cite{Goyal_Mahajan_Gupta_Misra_2019, Islam_Chen_Panda_Karlinsky_Radke_Feris_2021, Zhai_Puigcerver_Kolesnikov_Ruyssen_Riquelme_Lucic_Djolonga_Pinto_Neumann_Dosovitskiy_2020}. We believe SSFL is effective and robust for RSIs understanding tasks when the labeled data are insufficient. The reasons are:

\begin{itemize}
\item From the perspective of data: the rapidly evolving earth observation system provides abundant remote sensing data \cite{Long_Xia_Li_Yang_Yang_Zhu_Zhang_Li_2021, Steffen_Richardson_Rockstrom_Schellnhuber_Dube_Dutreuil_Lenton_Lubchenco_2020}. However, most of these data are unlabeled, so we cannot use them directly for SFL. Learning features from massive unlabeled data in a self-supervised manner can alleviate dependence on labeled samples as well as the sample-imbalance problem in SFL \cite{Liu_HaoChen_Gaidon_Ma_2021}.

\item  From the perspective of the feature learning mechanism: learning features in a label-free and task-independent way may be closer to the human visual process than SFL. The human visual recognition system is not limited to a specific task or specific dataset. Therefore, feature learning via supervised data-dependent and task-dependent ways may limit the generalization ability of feature representation. In addition, human language-based labels are not the prerequisite for constructing the human visual system. For example, a person who has no remote sensing knowledge can quickly extract the key features for distinguishing different land covers by observing a certain number of RSIs, and these features are invariant with the changes in lighting, perspective, and scale.
\end{itemize}

\section{Progress of self-supervised learning on remote sensing data}\label{sec:ssl}
Based on the unified framework of SSFL (Sec. \ref{sec:paradigm_ssl}), in this section,  we review research on SSFL in the remote sensing community following the proposed taxonomy: data, SSFL signals, and evaluation methods. Section \ref{sec:ssl_dataset} describes the progress in constructing large remote sensing datasets that can be used for SSFL. Section \ref{sec:ssl_signal} reviews how scholars construct SSFL signals to learn features from unlabeled remote sensing datasets. Section \ref{sec:ssl_evaluation} lists the common methods to evaluate the features learned by SSFL signals.

\subsection{Self-supervised learning datasets}\label{sec:ssl_dataset}
Studies \cite{Goyal_Caron_Lefaudeux_Xu_Wang_Pai_Singh_Liptchinsky_Misra_Joulin_2021, Jing_Tian_2021, Radford_Kim_Hallacy_Ramesh_Goh_Agarwal_Sastry_Askell_Mishkin_Clark_2021, Sun_Shrivastava_Singh_Gupta_2017, Yuan_Chen_Chen_Codella_Dai_Gao_Hu_Huang_Li_Li_2021} have shown that the models trained on large-scale and diverse datasets have two advantages: 1) The learned parameters provide a good starting point, so models can converge faster when they are trained on other tasks; 2) Such models usually have learned rich features, which help to reduce the over-fitting risk when they are trained on new tasks. From the perspective of data volume and semantic diversity, this section reviews the RSI datasets suitable for SSFL.

\begin{table*}[!t]
  \centering
  \renewcommand{\arraystretch}{1.2}
  \caption{The manually constructed large-scale remote sensing image datasets suitable for self-supervised learning.}
    \begin{tabular}{cccccccc}
    \toprule
    Dataset & Year & Categories & \#Num.  & Image size & Resolution (m) & Image Type & Image source \\
    \midrule
    RSD46-WHU & 2017 & 46 & 117,000 & 256 & 0.5 $\mathrm{\sim}$ 2 & RGB & Google Earth, Tianditu \\
    fMoW & 2018 & 63 & 1,047,691 & - & - & Multispectral & Digital Globe \\
    Million-AID & 2021 & 51 & 1,000,848 & 110 $\mathrm{\sim}$ 31,672 & 0.5 $\mathrm{\sim}$ 153 & RGB & Google Earth \\
    BigEarthNet & 2019 & 19 & 590,326 & 20 $\mathrm{\sim}$ 120 & 10 $\mathrm{\sim}$ 60 & Multispectral & Sentinel-2 \\
    SEN12MS & 2019 & 33 & 180,662 & 256 & 10 & SAR-Multispectral & Sentinel-1, Sentinel-2 \\
    SeasoNet & 2022 & 33 & 1,759,830 & 20 $\mathrm{\sim}$ 120 & 10 $\mathrm{\sim}$ 60 & Multispectral & Sentinel-2 \\
    \bottomrule
    \end{tabular}
  \label{tab:labeled_datasets}%
  \leftline{\qquad \qquad \#Num. is the number of samples}
\end{table*} 

\begin{figure*}[!t]
  \centering
  \includegraphics*[width=0.60\linewidth]{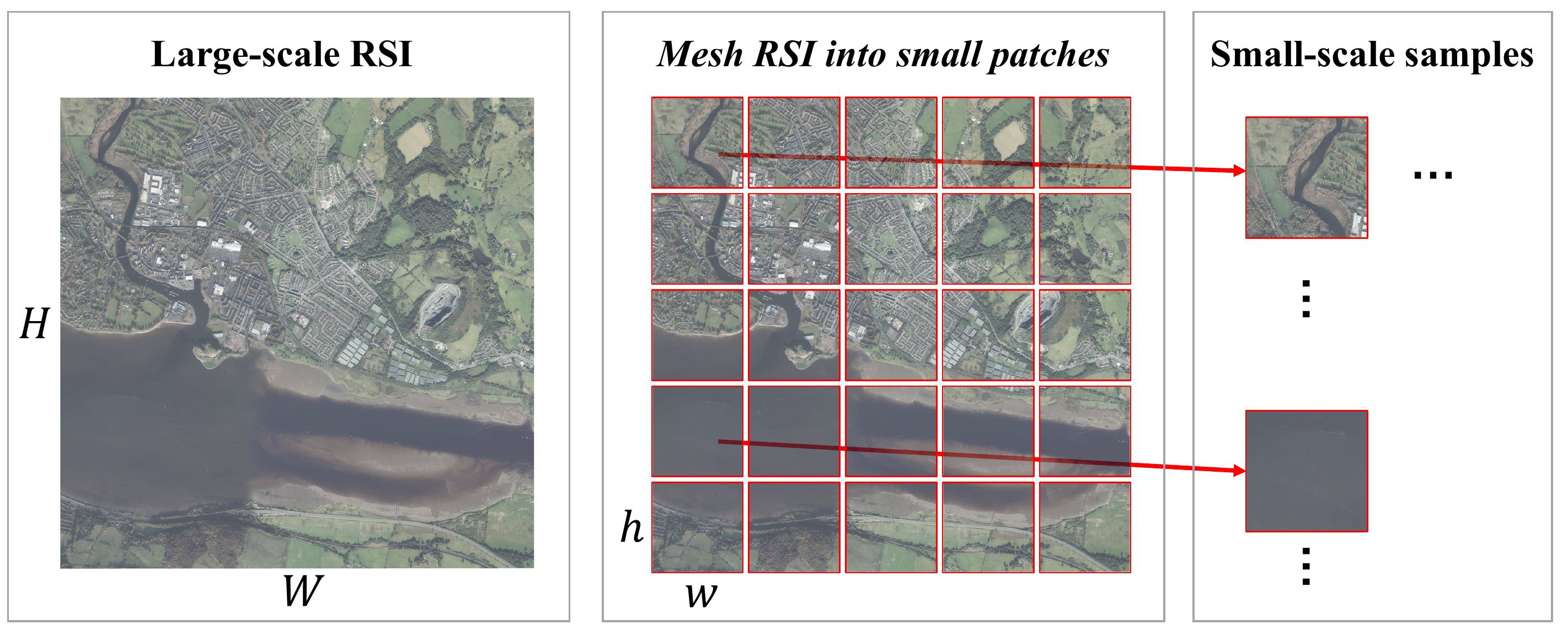}
  \caption{Illustration of the traditional grid sampling method.}
  \label{fig:grid_sampling}
\end{figure*}  

\subsubsection{Manually constructed labeled remote sensing image datasets}
In the past decades, researchers in the RS field have released many high-quality RSI datasets, as shown in Table \ref{tab:labeled_datasets}. RSD46-WHU \cite{Xiao_Long_Li_Wei_Tang_Liu_2017} is the first high-resolution RSI dataset with a sample size of more than 100,000, containing 117,000 labeled samples of 46 scene categories from Google Earth and Tianditu platforms. Later, Christie et al. \cite{Christie_Fendley_Wilson_Mukherjee_2018} constructed the Functional Map of the World (fMoW) dataset, consisting of 132,716 sample pairs. Each sample pair contains multi-temporal and multi-modal images collected from many satellites of the DigitalGlobe series, so fMoW has up to 1,047,691 medium- and high-resolution images. More importantly, the fMoW dataset spatially covers more than 200 countries, temporally spans 15 years, and has 62 types of RS scenes in terms of semantic content. Therefore, compared with other small-scale datasets, the fMoW dataset provides a more comprehensive representation of the real land cover, which allows models to learn diverse data features. Recently, Long et al. \cite{Long_Xia_Li_Yang_Yang_Zhu_Zhang_Li_2021, Long_Xia_Zhang_Cheng_Li_2022} constructed the first million-scale high-resolution RSI scene classification dataset in the RS field, called Million-AID. It contains 1,000,848 land cover and land use scenes of 51 categories. By collecting aerial images under various circumstances (i.e., illumination, viewpoint, scale, weather, season, geo-location), Million-AID can well represent the feature distribution of RS scenes in the real world.

For the low- and medium-resolution multispectral datasets, Sumbul et al. \cite{Sumbul_Charfuelan_Demir_Markl_2019} constructed a scene classification dataset named BigEarthNet, containing 590,326 samples collected from Sentinel-2 satellite data, and assigned one or more land cover labels to each sample. The SeasoNet \cite{Kobmann_Brack_Wilhelm_2022} dataset is also constructed based on Sentinel-2 satellite data, which has nearly two million samples with pixel-level labels. In addition, Schmitt et al. \cite{Schmitt_Hughes_Qiu_Zhu_2019} constructed a large-scale multimodality land cover classification dataset, SEN12MS, based on Sentinel-1 and Sentinel-2 satellite data, which contains 180,662 SAR-multispectral image pairs sampled from different regions and seasons and thus has a wide temporal and spatial coverage. The SEN12MS can be used for joint learning of multi-modal RS data features.

Self-supervised learning does not require manual annotation, so the above datasets can be directly used for SSFL after discarding labels. However, the size of these datasets is much smaller than that of the datasets used for computer vision, which is as large as hundreds of millions. Therefore, there is an urgent need to build larger datasets for SSFL.

\begin{table*}[!t]
  \centering
  \renewcommand{\arraystretch}{1.2}
  \caption{Remote sensing image dataset constructed by automated sampling.}
    \begin{tabular}{ccccccc}
    \toprule
    Dataset & Year & \#Num. & Image size & Resolution (m) & Image Type & Image source \\
    \midrule
    SoundingEarth & 2021 & 50,545 & 1024 & 0.2 & RGB - Audio & Google Earth \\
    SeCo & 2021 & 1,000,000 & - & 10 $\mathrm{\sim}$ 60 & Multispectral & Sentinel-2 \\
    TOV-RS-Balanced & 2022 & 500,000 & 600 & 1$\mathrm{\sim}$20 & RGB  & Google Earth \\
    SSL4EO-S12 & 2022 & 3,012,948 & 20 $\mathrm{\sim}$ 120 & 10 $\mathrm{\sim}$ 60 & SAR - Multispectral & Sentinel-1, Sentinel-2 \\
    \bottomrule
    \end{tabular}
  \label{tab:ssl_datasets}%
  \leftline{\qquad \qquad \qquad \#Num. is the number of samples}
\end{table*} 

\subsubsection{Remote sensing image dataset constructed by automated sampling}
By self-supervised learning, the feature learning process does not require manual labeling, which reduces the labeling cost for constructing ultra-large-scale RSI datasets.

Grid sampling is a traditional way to automatically collect RSI samples. As shown in Figure \ref{fig:grid_sampling}, the grid sampling method meshes a large-scale RSI of size $H \times W$ into $\frac{H}{h} \times \frac{W}{w}$ non-overlapped patches of size $h \times w$. This method can quickly and easily build a large-scale self-supervised learning dataset based on massive RSIs. However, since the spatial distribution of geographic elements is naturally unbalanced \cite{Gong_Liu_Zhang_Li_Wang_Huang_Clinton_Ji_Li_Bai_2019, Jun_Ban_Li_2014}, this method can hardly ensure data diversity.

To solve the problem, Manas et al. \cite{Manas_Lacoste_Giro-i-Nieto_Vazquez_Rodriguez_2021} collected Sentinel-2 images from urban areas globally to construct a large-scale dataset, called SeCo. This is based on the hypothesis that cities and their surrounding areas have the most diverse and relatively balanced land cover types. SeCo contains more than 1 million RSIs from over 2,000 urban areas in different seasons. Further, Wang et al. \cite{Wang_Braham_Albrecht_Xiong_Liu_Zhu_2022} constructed a self-supervised dataset, SSL4EO-S12, containing three million image samples, larger than all previous datasets. Compared with SeCo, SSL4EO-S12 covers a wider spatio-temporal range, and contains multi-modal remote sensing data including Sentinel-1 and Sentinel-2, supporting multimodal remote sensing feature learning. In addition, to enable the model to learn image-audio features, Heidler et al. \cite{Heidler_Mou_Hu_Jin_Li_Gan_Wen_Zhu_2021} constructed the first large-scale RSI-audio dataset, called SoundingEarth, by two steps: 1) Crawling surface audio data with geo-coordinates from an open source project called Radio Aporee:::Maps\footnote{https://archive.org/details/radio-aporee-maps}; 2) Collecting the corresponding RSIs from Google Earth platform based on the geo-coordinates of the audio. SoundingEarth contains about 3,500 hours of audio data and 50,545 RSIs with a spatial resolution of approximately 0.2 m/pixel.

In addition to semantic diversity, the class balance and image resolution of the self-supervised learning datasets are also crucial for the model to learn valuable images \cite{Tao_Qi_Zhang_Zhu_Lu_Li_2022}. Therefore, Tao et al. proposed an automatic RSI sampling method guided by geographic data products. To guarantee semantic diversity and image resolution, RSIs with a resolution of 1m/pixel are collected from the Google Earth platform. Then, the label from a global land cover product FROM\_GLC10 \cite{Gong_Liu_Zhang_Li_Wang_Huang_Clinton_Ji_Li_Bai_2019} is used to guide the sampling of natural geographical samples (e.g., woodlands and grasslands). In addition, this method uses the label from Open Street Map (OSM) to guide the sampling of man-made geographical elements (e.g., airports, parking lots, and schools). In this way, they first obtained a dataset containing more than 3 million RSIs, called the TOV-RS dataset, with a balanced number of natural and man-made samples. Further, to ensure a balanced sampling of the subcategories of natural and man-made categories, a class-balanced resampling strategy is proposed to post-process the obtained samples by their labels.  Finally, they obtain a relatively class-balanced dataset, called the TOV-RS-balanced dataset. Table \ref{tab:ssl_datasets} summarizes the basic characteristics of the existing automated sampling-based datasets.

\begin{figure*}[!t]
  \centering
  \includegraphics*[width=0.72\linewidth]{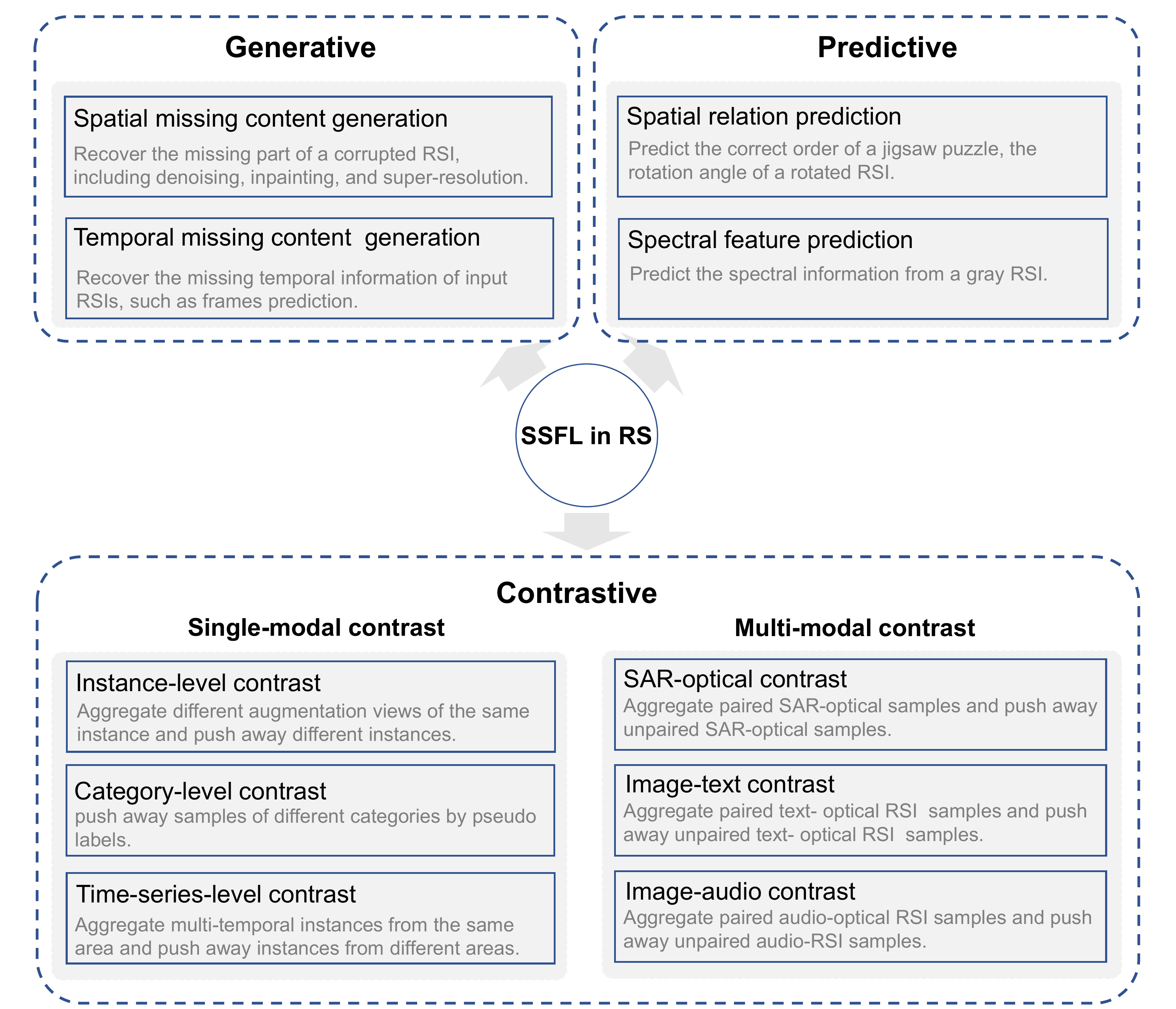}
  \caption{Categories of the self-supervised feature learning signal in the remote sensing field: generative, predictive, and contrastive.}
  \label{fig:ssl_overview}
\end{figure*}  

\subsection{Self-supervised feature learning signals}\label{sec:ssl_signal}

In SSFL, feature learning does not rely on manual labels but is guided by pseudo labels, which are obtained automatically by mining the association information from massive unlabeled data with human-designed self-learning signals. Many studies showed that the choice of SSFL signal is crucial for the model learning ability of good features \cite{Ericsson_Gouk_Hospedales_2021, Goyal_Mahajan_Gupta_Misra_2019, Kolesnikov_Zhai_Beyer_2019, Newell_Deng_2020, Tao_Qi_Lu_Wang_Li_2022, Zhai_Puigcerver_Kolesnikov_Ruyssen_Riquelme_Lucic_Djolonga_Pinto_Neumann_Dosovitskiy_2020}. As shown in Figure \ref{fig:ssl_overview}, the existing SSFL signals can be classified into three categories: generative, predictive, and contrastive. In this section, the principles for designing these signals and the characteristics of the learned features are introduced.

\subsubsection{Generative learning signals}

Generative learning signals train the model to reconstruct an original input from a partially corrupted one for feature learning. It assumes that the model can recover the missing information if the contextual information features are well-learned. The construction process of this learning signal is as follows:

Step 1: Corrupting the origin data $\boldsymbol{x}$ by adding random noise, masks, or down-sampling $\boldsymbol{x}$ to obtain a destroyed version $\widetilde{\boldsymbol{x}}$.

Step 2: A model $f(\cdot)$ with an encoder-decoder architecture learns features by minimizing the objective function $||{f(\widetilde{\boldsymbol{x}}) - \boldsymbol{x}}||^2_2$.

The generative learning signals can be further divided into learning signals based on spatial missing content generation and that based on temporal missing content generation.

\paragraph{\textbf{Learning signals based on spatial missing content generation}}
\begin{itemize}
  \item \textbf{Image-denoising}: Since noise leads to image content missing, early studies construct this learning signal using image-denoising tasks for learning abstract and robust features \cite{Vincent_Larochelle_Lajoie_Bengio_Manzagol_2010, Vincent_Larochelle_Bengio_Manzagol_2008}. These studies are commonly based on the hypothesis that the model must learn stable spatial features to recover clear images from the noise images. Based on the hypothesis, image denoising-based SSFL methods have been used to learn features from various RSIs \cite{Molini_Valsesia_Fracastoro_Magli_2022, Yuan_Guan_Sun_2019, Imamura_Itasaka_Okuda_2019, Qian_Zhu_Chen_Zhou_2022, Wang_Luo_Li_Hu_Zhang_Zhong_2022}. For example, Zhang et al. \cite{Zhang_Gong_Su_Liu_Li_2016} employ a stacked auto-encoder to learn spatial features from both SAR and high-resolution optical RSIs via the denoising task and thus benefit downstream cross-modal change detection tasks.
  \item \textbf{Inpainting}: Pathak at el. further developed a similar learning signal using a more challenging inpainting task \cite{Pathak_Krahenbuhl_Donahue_Darrell_Efros_2016}, which corrupts images by patch-level masks instead of pixel-level noise. The motivation is that the model should capture higher-level context features to predict a reasonable hypothesis for the missing part(s) of the input image. Similar work can be found in \cite{Pathak_Krahenbuhl_Donahue_Darrell_Efros_2016, Li_Chen_Shi_2021, Iizuka_Simo-Serra_Ishikawa_2017, Xue_Yu_Yu_Liu_Zhang_Wu_2022}. Singh et al. \cite{Singh_Batra_Pang_Torresani_Basu_Paluri_Jawahar_2018} used this kind of SSFL signal to learn spatial features from high-resolution RSIs. To increase the inpainting task difficulty, they introduced an adversarial training framework to select texturally complex regions for masking instead of masking randomly selected regions. Their experiments showed that increasing the difficulty of inpainting tasks can improve the model learning ability of better features for downstream semantic segmentation tasks. Similarly, the masked auto-encoder (MAE) method makes the inpainting task more challenging by masking more than 75\% of the image \cite{He_Chen_Xie_Li_Dollar_Girshick_2022}, thus stimulating the learned transformer model \cite{Dosovitskiy_Beyer_Kolesnikov_Weissenborn_Zhai_Unterthiner_Dehghani_Minderer_Heigold_Gelly_2021, Liu_Lin_Cao_Hu_Wei_Zhang_Lin_Guo_2021} to capture key features of images. Most recently, MAE-based SSFL methods have received much attention for their power of generic feature learning \cite{Chen_Ding_Wang_Xin_Mo_Wang_Han_Luo_Zeng_Wang_2022, Gao_Ma_Li_Lin_Dai_Qiao_2022, Shi_Siddharth_Torr_Kosiorek_2022, Wei_Fan_Xie_Wu_Yuille_Feichtenhofer_2022, Xie_Zhang_Cao_Lin_Bao_Yao_Dai_Hu_2022}. Sun et al. \cite{Sun_Wang_Lu_Zhu_Lu_He_Li_Rong_Yang_Chang_2022} applied the MAE method to SSFL for a 3 million unlabeled RSI dataset. Experiments show that the learned features generalize well to various downstream tasks, including scene classification, target recognition, semantic segmentation, and change detection.
\begin{figure}[!t]
  \centering
  \includegraphics*[width=0.99\linewidth]{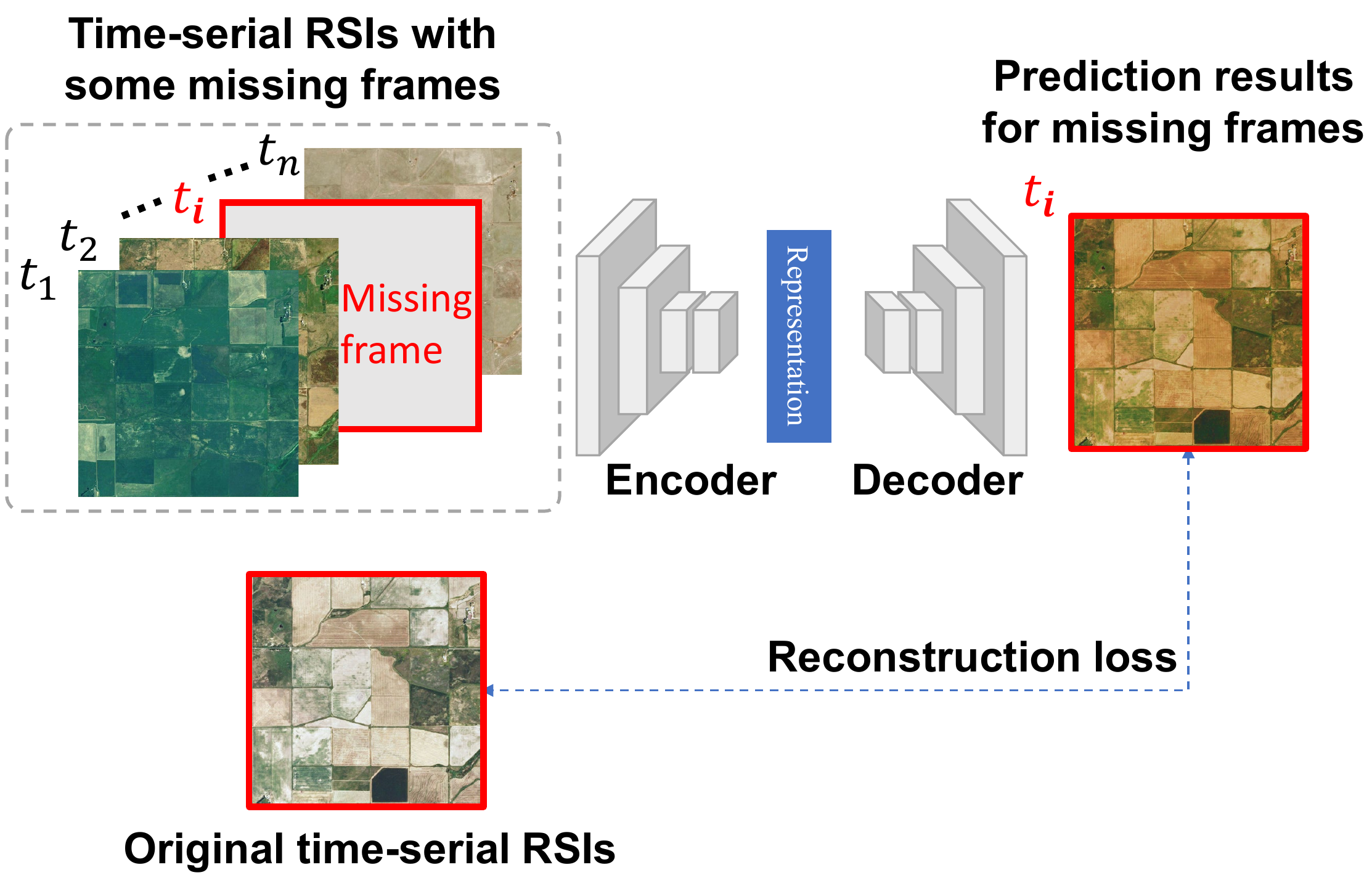}
  \caption{Illustration of the temporal missing content generation.}
  \label{fig:ssl_generative_temporal}
\end{figure}  
  \item \textbf{Super-resolution}: The super-resolution-based SSFL signal follows the above idea of image missing content generation to recover high-resolution images from blurred low-resolution images for capturing spatial features of the various object profiles \cite{Ledig_Theis_Huszar_Caballero_Cunningham_Acosta_Aitken_Tejani_Totz_Wang_2017, Liu_Liu_Hou_Tao_Han_2021, Nguyen_Anger_Davy_Arias_Facciolo_2021}. For paired hyper-spectral and multi-spectral images, Vedaldi et al. \cite{Yao_Hong_Chanussot_Meng_Zhu_Xu_2020} proposed a two-branch super-resolution model for learning coupled spatial-spectral features based on the hypothesis of content consistency between the two modal data. However, the generalization performance of the features learned by the learning signal based on the super-resolution tasks in RSI understanding tasks remains unexplored.
\end{itemize}

\begin{figure*}[!t]
  \centering
  \includegraphics*[width=0.68\linewidth]{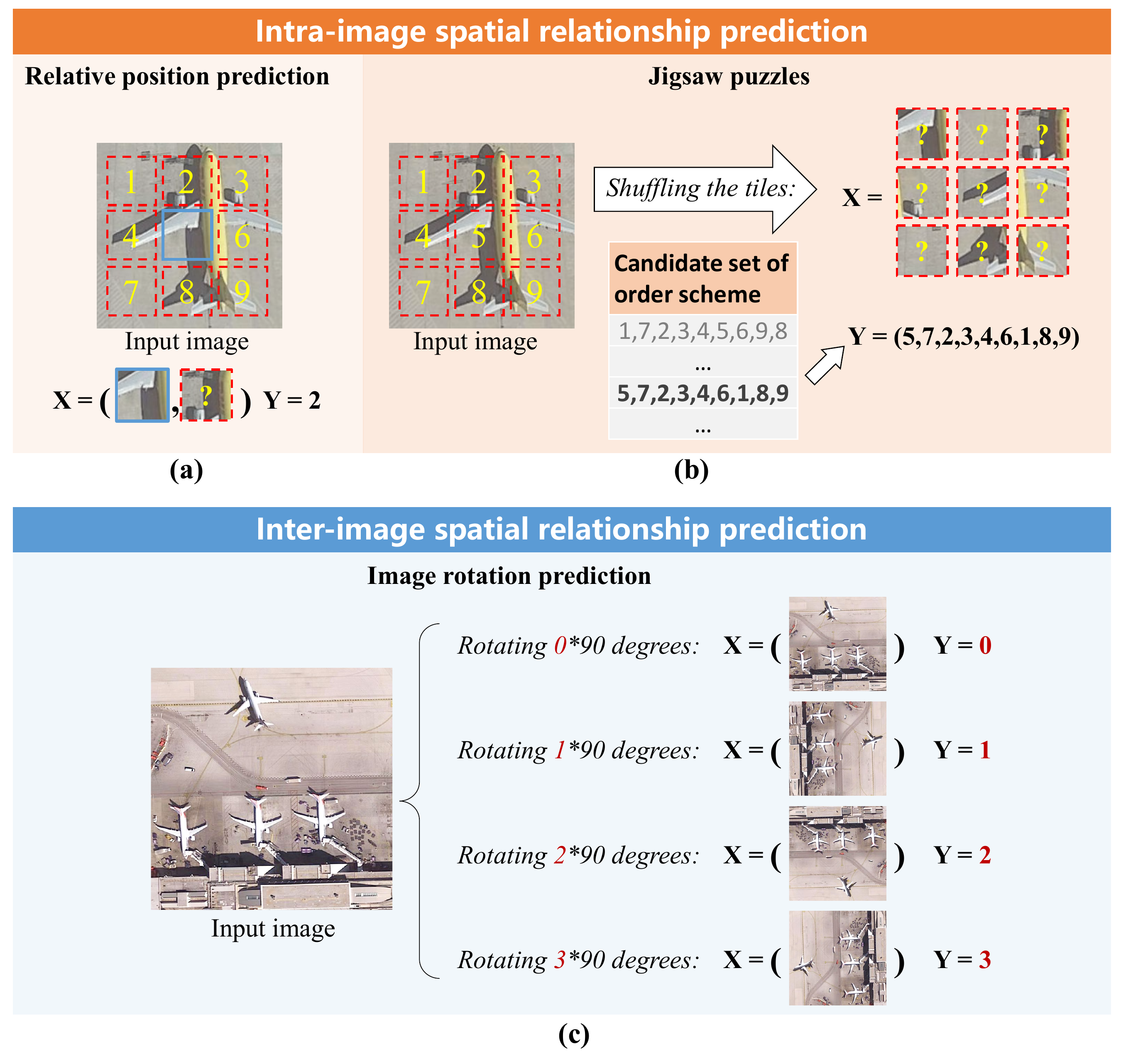}
  \caption{Illustration of three types of spatial relationship prediction tasks: (a) relative position prediction, (b) jigsaw puzzle, and (c) image rotation angle prediction. (a) and (b) aim at predicting intra-image spatial relationships. (c) aims at predicting inter-image spatial relationships.}
  \label{fig:ssl_predictive_spatial}
\end{figure*}  

\paragraph{\textbf{Learning signals based on temporal missing content generation}}
Constructing such a learning signal for SSFL by temporal missing content generation tasks is based on the assumption that there exist correlations between temporal variation patterns and semantics \cite{Srivastava_Mansimov_Salakhutdinov_2015, Finn_Goodfellow_Levine_2016, Villegas_Yang_Hong_Lin_Lee_2018, Kumar_Tripathi_Pant_2021, Peng_Huang_Vongkusolkit_Gao_Wright_Fang_Qiang_2021, Yuan_Lin_2021, Feichtenhofer_Fan_Li_He_2022, Yuan_Lin_Liu_Hang_Zhou_2022}. For example, as shown in Figure \ref{fig:ssl_generative_temporal}, a model with an encoder-decoder structure should exploit the learned temporal pattern of land cover to reconstruct the missing frame of the input time-series data. Yuan and Lin \cite{Yuan_Lin_2021} applied this idea to time-series Sentinel-2 images to learn spectral-temporal features related to crop semantics, as the phenological change pattern of crops on time-series RSIs can reflect the crop type. They add random noise to the original time-series images and construct a transformer model to learn features by recovering clean versions. To increase the difficulty of temporal missing content generation tasks, Yuan et al. \cite{Yuan_Lin_Liu_Hang_Zhou_2022} proposed a patch-level generative SSFL method that randomly masks some frames from the time-series Sentinel-2 images completely rather than adding only pixel-level noise. Using the same datasets as \cite{Yuan_Lin_2021}, their experiments show that the method can further improve the performance of the pre-trained transformer in crop classification.

\subsubsection{Predictive learning signals}\label{sec:ssl_signal_predictive}
Unlike the generative learning signals that deal with pixel-level details, the predictive learning signal focuses on learning semantics context features. Such learning signals can be divided into two categories: learning signals based on spatial relation prediction and that based on spectral feature prediction.

\paragraph{\textbf{Learning signals based on spatial relation prediction}}\label{sec:ssl_signal_predictive_spatial}
The application of this kind of signal in SSFL is based on the assumption that the spatial relation information between object parts is correlated with object semantics. For example, an airplane is composed of many parts including wings, a fuselage, tailplane in a fixed spatial combination. This type of signal is constructed in the following common ways:
\begin{itemize}
  \item \textbf{Relative position prediction}: Doersch et al., \cite{Doersch_Gupta_Efros_2015} forced the SSFL model to predict the relative positions of two image tiles cropping from the input image. As shown in Figure \ref{fig:ssl_predictive_spatial}(a), one tile is in the middle and the other tile is randomly located at any position around the former. Their method is based on a hypothesis that the model can understand what the object is in order to accurately predict the spatial relations between the parts of the object.
  \item \textbf{Jigsaw puzzle}: Based on the same hypothesis \cite{Doersch_Gupta_Efros_2015}, the jigsaw puzzles task \cite{Noroozi_Favaro_2016, Cruz_Fernando_Cherian_Gould_2019, Kim_Cho_Yoo_Kweon_2018, Noroozi_Vinjimoor_Favaro_Pirsiavash_2018, Wei_Xie_Ren_Xia_Su_Liu_Tian_Yuille_2019, Haghighi_Hosseinzadeh_Taher_Zhou_Gotway_Liang_2020, Chen_Liu_Jia_2021} is also used to construct the learning signal for SSFL. For example, Noroozi and Favaro \cite{Noroozi_Favaro_2016} constructed a jigsaw puzzle prediction task by randomly shuffling tiles of an image as shown in Figure \ref{fig:ssl_predictive_spatial}(b). The SSFL model is then trained to learn semantic context features by selecting the correct one from the candidate tile order schemes.
\end{itemize}

There are two key challenges \cite{Cruz_Fernando_Cherian_Gould_2019, Doersch_Gupta_Efros_2015, Noroozi_Favaro_2016} in learning the desired spatial context features for the learning signals constructed in the ways described above:
 i) Shortcuts: Texture and color continuity between image tiles is simple cues that can serve as shortcuts for models to solve the above prediction tasks. Unfortunately, shortcuts undermine the model’s motivation to learn desired features, reducing the performance of the learned features in downstream tasks. A possible solution to this problem is to use harder versions of the spatial relation prediction task or integrate multiple tasks to construct learning signals. For example, Kim et al. \cite{Kim_Cho_Yoo_Kweon_2018} created a complex composite task by integrating the jigsaw puzzle with colorization and inpainting tasks. In this composite task, the model has to solve a more difficult version jigsaw puzzle, in which one image tile is dropped and the rest are decolored. Then, the model is trained to reconstruct the missing image tile and predict the color of the rest ones.
 ii) Task ambiguities. Ambiguity is common in jigsaw puzzles when image tiles lack valid content or are highly similar to each other \cite{Cruz_Fernando_Cherian_Gould_2019, Noroozi_Favaro_2016}. In these cases, SSFL models can only solve the tasks by guessing or using other shortcuts rather than by learning meaningful spatial contextual features. This problem is especially common for RSIs of homogenized scenes such as sea, deserts, and forests. The above points were confirmed by the experiments of \cite{Tao_Qi_Lu_Wang_Li_2022}, which found that jigsaw puzzles cannot achieve satisfactory performance in the downstream task of high-resolution RSI scene classification.

In addition to using intra-image spatial relationships, inter-image spatial relationships can also be exploited to construct the learning signal:
\begin{itemize}
 \item \textbf{Image rotation prediction}: Using the image rotation prediction task to construct the signal \cite{Gidaris_Singh_Komodakis_2018, Feng_Xu_Tao_2019, Chen_Zhai_Ritter_Lucic_Houlsby_2019, Yun_Park_Cho_2020, Li_Hu_Qi_Yu_Zhao_Heng_Xing_2021} is based on the assumption that an SSFL model that can accurately predict the rotation angle of the input image, should understand the concept of objects in the image (such as the position, type, and pose of the aircraft in Figure \ref{fig:ssl_predictive_spatial} (c)). The construction process of this learning signal is as follows (Figure \ref{fig:ssl_predictive_spatial}(c)):

Step 1: Given a predefined set of discrete rotation transformations $G={\{g(\cdot | y)\}}^K_{y=0}\ $, the rotated image $\widetilde{\boldsymbol{x}}$ is obtained by applying a random rotation transformation $g(\cdot | y)$ from $G$ to the input image $\boldsymbol{x}$. whose corresponding label is $y$. $g(\cdot | y)$ means rotating the image by $y$*90 degrees.

Step 2: Using ($\widetilde{\boldsymbol{x}}$, $y$) as the supervision for training the SSFL model.

In the above steps, the definition of G affects the performance of the learned features in downstream tasks. Empirical experiments of \cite{Gidaris_Singh_Komodakis_2018} found that a small number of rotation angles (e.g., K=2) may lead to insufficient monitoring information, while a larger K (e.g., K=8) leads to task ambiguity. Zhao et al. \cite{Zhao_Luo_Li_Chen_Piao_2020} performed SSFL on high-resolution optical RSIs by this learning signal and visualized the feature maps of scenes like airports and sea surfaces. The results show that unlike using supervised learning only, the model learned via image rotation prediction will pay more attention to key objects in images which helps to improve the accuracy of RSI scene classification. Some work used this learning signal to exploit high-resolution \cite{Dang_Li_2021, Ji_Gao_Zhang_Wan_Li_Mei_2022} or hyperspectral RSI data \cite{Paoletti_Haut_Roy_Hendrix_2020, Yue_Fang_Rahmani_Ghamisi_2022} for SSFL, thus benefiting downstream interpretation tasks based on the corresponding data. Further studies \cite{Dangovski_Jing_Loh_Han_Srivastava_Cheung_Agrawal_Soljacic_2022, Lee_Kim_Cho_2022} have found that rotation-equivariant features \cite{Qi_Zhang_Chen_Tian_2019} varying accordingly with the rotation of the objects (e.g., aircraft, ship, buildings) could be learned by this signal. And Zhang et al. \cite{Zhang_Wen_Liu_Pan_2019} suggested that the rotation-equivariant feature learned from this SSFL signal has better performance in object detection tasks on SAR RSIs than traditional CNNs that focus only on rotation-invariant features. Recent studies \cite{Wen_Liu_Zhang_Pan_2021, Xu_Cui_Guo_Zhang_Yu_2021} also confirmed that learning rotation-equivariant features in a self-supervised manner reduces the dependence of the object detection task on the labeled sample volume. However, this learning signal is not favored in rotation agnostic RSIs like the domed buildings, oceans, and dessert.
\end{itemize}

\begin{figure}[!t]
  \centering
  \includegraphics*[width=0.95\linewidth]{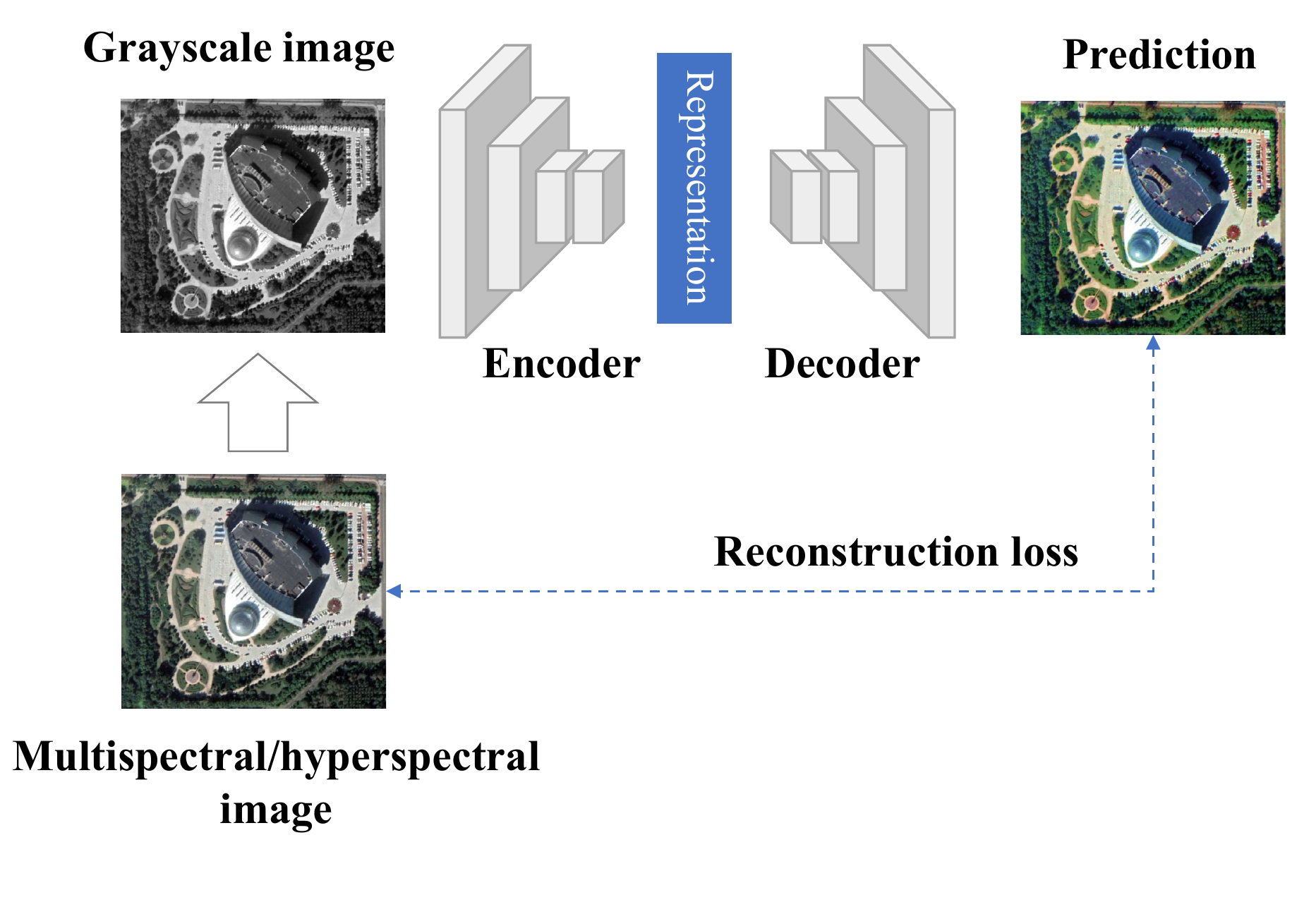}
  \caption{Illustration of spectral feature prediction.}
  \label{fig:ssl_predictive_spectral}
\end{figure}  

\paragraph{\textbf{Learning signals based on spectral feature prediction}}
Using this self-supervised learning signal for feature learning is based on the assumption of a strong correlation between semantics and its corresponding spectra \cite{Zhang_Isola_Efros_2016, Larsson_Maire_Shakhnarovich_2016, Zhang_Isola_Efros_2017, Larsson_Maire_Shakhnarovich_2017, Stojnic_Risojevic_2018b}. For example, vegetation preferentially reflects more near-infrared and green light than other wavelengths of light, so it appears green. And the seawater regions are blue because water preferentially absorbs red spectra. Therefore, the model should understand the semantics of the image to correctly predict the corresponding spectrum. The construction process of this learning signal is as follows (Figure \ref{fig:ssl_predictive_spectral}):

Step 1: Obtaining gray-scale image $\widetilde{\boldsymbol{x}}$ from an input multispectral/hyperspectral image $\boldsymbol{x}$.

Step 2: Training a model $f(\cdot )$ with encoder-decoder structure by minimizing the object function $||{f(\widetilde{\boldsymbol{x}})-\boldsymbol{x}}||^2_2$.

\begin{figure*}[!t]
  \centering
  \includegraphics*[width=0.87\linewidth]{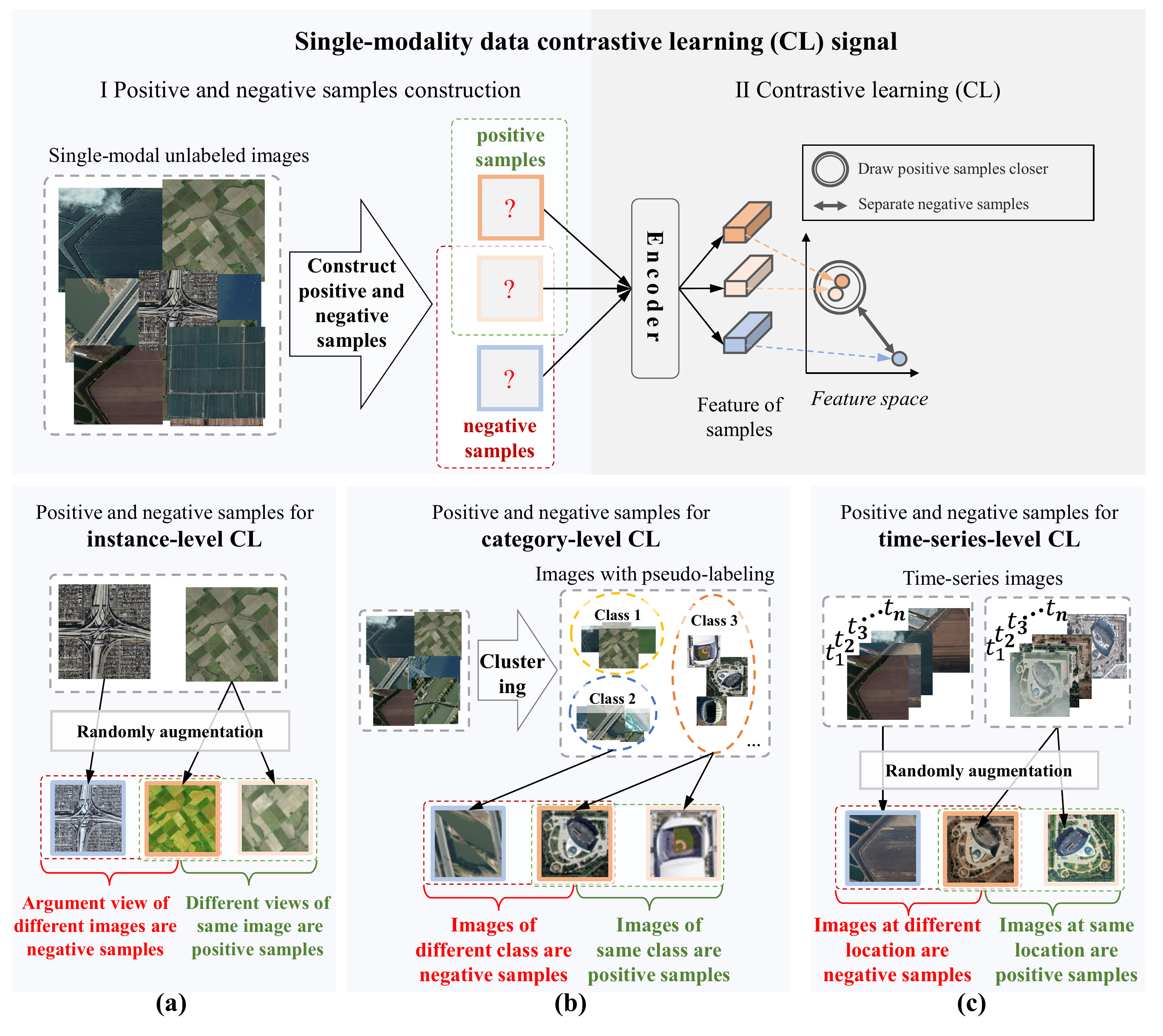}
  \caption{Illustration of three kinds of single-modal contrastive learning signals: instance-level, category-level, and time-series-level.}
  \label{fig:ssl_contrastive_single_modal}
\end{figure*}  

For RSI with only RGB channels, the above process can be regarded as a colorization task, i.e., predicting the RGB color channels using the gray-scale channels. Zhang et al. \cite{Zhang_Isola_Efros_2016} first applied this idea to construct the learning signal for SSFL. They also considered the shortcuts and task ambiguities problems. But for the learning signals based on spectral feature prediction, these two problems are caused by reasons different from those mentioned in section \ref{sec:ssl_signal_predictive_spatial}. They suggest constructing the colorization task in Lab space instead of RGB color space because the gray-scale channel L and the color channels ab are independent. Thus, to complete the task, the model should understand image semantics rather than use the inter-channel correlation as a shortcut. Empirical experiments \cite{Stojnic_Risojevic_2018a} show that the learned features using Lab space are more beneficial for downstream RSI scene classification tasks than using RGB space. Task ambiguity may occur in the colorization task because for both natural and remote sensing images, different objects share the same spectrum and one object may have different spectra. In such cases, the model completes the colorization task by guessing rather than understanding the semantics of the image. Therefore, Zhang et al. \cite{Zhang_Isola_Efros_2016} transformed the color value regression problem in colorization tasks into a classification problem to select the most suitable color scheme. Also to reduce task ambiguity, Larsson et al. \cite{Larsson_Maire_Shakhnarovich_2016} constructed a learning signal using spectral histogram prediction tasks instead of spectral value regression. For multispectral RSIs, Vincenzi et al. \cite{Vincenzi_Porrello_Buzzega_Cipriano_Fronte_Cuccu_Ippoliti_Conte_Calderara_2021} suggested training the model by predicting color channels based on multi-spectral channels not including RGB instead of gray-scale channels, to fully exploit the diverse features of each band.

\subsubsection{Contrastive learning signals}\label{sec:ssl_signal_contrastive}

Extensive experimental work in psychology has shown that infants learn perceptual categories primarily through observation rather than linguistic supervision \cite{Orhan_Gupta_Lake_2020}. In this way, they can summarize different manifestations of the same object (i.e., invariance) and distinguish objects by different manifestations (i.e., distinguishability). Contrast learning signals are designed to mimic this process, which brings different augmented views (positive sample pairs) of the same image closer and separate views (negative sample pairs) of different images, to learn both invariant and distinguishable visual features \cite{Orhan_Gupta_Lake_2020, Jaiswal_Babu_Zadeh_Banerjee_Makedon_2021}. The contrastive learning signals are constructed by the following two steps:

Step 1: Given an unlabeled training set $X=\{{\boldsymbol{x}}_1,\ {\boldsymbol{x}}_2,\dots ,\ {\boldsymbol{x}}_n\}$, each sample ${\boldsymbol{x}}_i$ is augmented by $T(\cdot )$ to create a pair of positive samples (${\boldsymbol{x}}^1_i, {\boldsymbol{x}}^2_i$). In contrast, any two augmented views of different samples are treated as a negative sample pair $({\boldsymbol{x}}^1_i\boldsymbol{,\ }{\boldsymbol{x}}^2_j$). Here, $T(\cdot )$ is a stochastic set of augmentations including random crop, random flip, color distortion, adding noise, and other common digital image processing operations.

Step 2: Training a model $f(\cdot )$ to discriminate the positive and negative samples by embedding them to a proper feature space using the loss function defined as Eq. \eqref{equ:equ_sscl}, where $\mathrm{sim}(\cdot, \cdot )$ indicates the similarity of the two sample feature vectors, generally using the cosine similarity. According to the type of data used, contrastive learning signals can be divided into two categories: single-modal contrastive learning signals and multi-modal contrastive learning signals.

\paragraph{\textbf{Single-modal contrastive learning signal}}\label{sec:ssl_signal_contrastive_single_modal}
Single-modal means that positive and negative samples used to construct this learning signal are collected from the same modality. Empirical experiments demonstrate that the way of constructing positive and negative samples has a significant effect on the performance of learned features on downstream tasks \cite{Huang_Yi_Zhao_2022, Tian_Sun_Poole_Krishnan_Schmid_Isola_2020, Wang_Isola_2020}. Thus, according to the construction ways of positive and negative samples, single-modal contrastive learning signals can be further classified into instance-level, category-level, and time-series-level signals, as shown in Figure \ref{fig:ssl_contrastive_single_modal}.

\begin{figure*}[hb]
\centering
\begin{equation}\label{equ:equ_sscl}
\mathcal{L}=-\mathrm{E}_X\left[\log \frac{\operatorname{sim}\left(f\left(\boldsymbol{x}_i^1\right)^T f\left(\boldsymbol{x}_i^2\right)\right)}{\operatorname{sim}\left(f\left(\boldsymbol{x}_i^1\right)^T f\left(\boldsymbol{x}_i^2\right)\right)+\sum_{j=1}^{n-1} \operatorname{sim}\left(f\left(\boldsymbol{x}_i^1\right)^T f\left(\boldsymbol{x}_j^2\right)\right)}\right]
\end{equation}
\end{figure*}

\textbf{Instance-level contrastive learning signal} \cite{Chopra_Hadsell_LeCun_2005, Weinberger_Saul_2009, Schroff_Kalenichenko_Philbin_2015, Oord_Li_Vinyals_2018, Wu_Xiong_Yu_Lin_2018, Hjelm_Fedorov_Lavoie-Marchildon_Grewal_Bachman_Trischler_Bengio_2019, Chen_Fan_Girshick_He_2020, Chen_Kornblith_Swersky_Norouzi_Hinton_2020, Bachman_Hjelm_Buchwalter_2019} takes different augmented views of the same image as positive samples and the views of any other images as negative samples. As a milestone work, Chen et al. \cite{Chen_Kornblith_Norouzi_Hinton_2020} proposed a simple yet powerful contrastive learning framework, called SimCLR. For constructing positive sample pairs, they claimed that the combination of multiple augmentation methods significantly outperformed a single data augmentation method. The reason is that the invariance of the features learned by this learning signal depends on the diversity of data augmentations applied to positive samples \cite{Bachman_Hjelm_Buchwalter_2019, Tian_Sun_Poole_Krishnan_Schmid_Isola_2020}. For negative sample construction, they found that a sufficiently large number of negative samples is crucial for the model to learn distinguishable features with high generalization. Their experiments quantitatively demonstrated that larger batch size (i.e., equal to the number of negative samples) and longer training cycles resulted in higher performance of the learned features. To learn features from more negative samples, He et al. \cite{He_Fan_Wu_Xie_Girshick_2020} proposed MoCo, which uses a dynamic queue to preserve the features of historical negative samples to obtain negative samples with a number far beyond the training batch size limit (i.e., generally equivalent to the number of negative samples). Kalantidis et al. \cite{Kalantidis_Sariyildiz_Pion_Weinzaepfel_Larlus_2020} further argued that increasing the number of negative samples alone does not guarantee performance improvement of contrastive learning. Therefore, they proposed to obtain hard negative samples by mixing the features of negative samples to learn more distinguishable features. From the perspective of model optimization, contrast learning between negative samples can avoid the trivial solution, i.e., the outputs of the model are always constant values. However, some novel methods\cite{Chen_He_2021, Grill_Strub_Altche_Tallec_Richemond_Buchatskaya_Doersch_Pires_Guo_Azar_2020, Marsocci_Scardapane_2022, Zbontar_Jing_Misra_LeCun_Deny_2021}, such as BYOL \cite{Grill_Strub_Altche_Tallec_Richemond_Buchatskaya_Doersch_Pires_Guo_Azar_2020} and SimSiam \cite{Chen_He_2021}, can perform SSFL with only positive sample pairs constructed and the learned features perform surprisingly well in various downstream vision tasks. These methods design the Siamese model with an online and an offline encoder in the same architecture and update the parameters of the two encoders separately. This avoids model performance collapse resulting from the absence of negative samples \cite{Tian_Chen_Ganguli_2021}.

In the field of RS, Tao et al. \cite{Tao_Qi_Lu_Wang_Li_2022} first applied the instance-level contrastive self-supervised method (i.e., SimCLR) to a high-resolution RSI scene interpretation task. They found that the self-supervised learning features outperformed the supervised learning features in scene classification tasks using only 10\% of the labeled samples required by the supervised method. On the basis of the first law of geography, Jean et al. \cite{Jean_Wang_Samar_Azzari_Lobell_Ermon_2019} assumed that geographically closer regions are more likely to have the same land cover or land use type. To construct the instance-level contrastive learning signal based on this assumption, they took two patches sampled from adjacent locations from a large-scale RSI as a positive sample pair, and another patch sampled from distant locations as the negative sample. Following the above paradigm, similar works are proposed for learning features from unlabeled RSIs for scene classification \cite{Jung_Jeon_2021, Kang_Fernandez-Beltran_Duan_Liu_Plaza_2021}.
In addition, considering the gap between instance-level contrastive learning and pixel-level segmentation, Li et al. \cite{Li_Li_Zhang_Liu_Huang_Zhu_Tao_2022} constructed a global style and local matching contrastive learning network (GLCNet) to extend instance-level contrast learning to super-pixel-level contrastive learning. In this network, different global view representations of the same image are taken as positive samples. Local views are randomly cropped from the positive sample pairs as additional positive samples for local detail feature learning. For the same purpose, some recent research \cite{Yang_Jiao_Liu_Hou_Yang_Zhang_Wang_2022, Zhu_Fan_Yang_Chen_2022} pays attention to the construction of fine contrast-level learning signals, e.g., pixel-level. For example, Muhtar et al. \cite{Muhtar_Zhang_Xiao_2022} introduced a pixel-level contrastive learning branch called index contrast, which first tracks the spatial index of each identical pixel across two views of the same image, and then takes the corresponding pixels of different views as additional positive samples. Experiments showed that the method outperformed the traditional instance-level contrast learning in land use classification tasks for four high-resolution RSIs because it can learn pixel-level features that are more suitable for semantic segmentation tasks. Similarly, scholars \cite{Xie_Zhan_Liu_Ong_Loy_2021} in computer vision are concerned about the gap between object-level recognition and this learning signal that aims at discriminating images. This is an important but unexplored issue for contrastive learning using RSIs that often cover complex scenes.

\textbf{Category-level contrastive learning signal}. Instance-level contrastive learning signals have been widely used for RSI feature learning \cite{Agastya_Ghebremusse_Anderson_Reed_Vahabi_Todeschini_2021, Guo_Xia_Luo_2021, Ren_Zhao_Hou_Chanussot_Jiao_2021, Seneviratne_Nice_Wijnands_Stevenson_Thompson_2021, Stojnic_Risojevic_2021, Xu_Sun_Chen_Lei_Ji_Kuang_2021, Zheng_Kellenberger_Gong_Hajnsek_Tuia_2021, Duan_Xie_Kang_Li_2022, Marsocci_Scardapane_Komodakis_2021, Jung_Oh_Jeong_Lee_Jeon_2022, Patel_Sharma_Pasquarella_Gulshan_2022, Scheibenreif_Hanna_Mommert_Borth_2022, Wang_Zhang_Han_2022}, but images with the same semantic content may be treated as negative sample pairs, which may mislead feature learning. This is called the false-negative sample problem in contrast learning \cite{Yang_Li_Huang_Liu_Hu_Peng_2021}.To address this problem, category-level contrastive learning signal only takes images of different “categories” as negative samples \cite{Bautista_Sanakoyeu_Tikhoncheva_Ommer_2016, Yang_Parikh_Batra_2016, Xie_Girshick_Farhadi_2016, Li_Cai_Zhang_Cai_Liu_2022, Ren_Yu_Wang_Liu_Zou_Wang_2022} rather than discriminating all images even if some of them have similar semantics. As no labels are available as category priors, this signal is constructed through the following steps (Figure \ref{fig:ssl_contrastive_single_modal}(b)):

Step 1: Samples are clustered by unsupervised approaches (e.g., k-means) to obtain pseudo-labels.

Step 2: According to pseudo-labels, samples in the same category are treated as positive samples, and those in different categories are treated as negative samples. By doing so, the false-negative sample problem is avoided.

As a milestone work, Caron et al. \cite{Caron_Bojanowski_Joulin_Douze_2018} proposed DeepCluster train the SSFL model by iterating the above two steps. Studies \cite{Van_Gansbeke_Vandenhende_Georgoulis_Proesmans_Van_Gool_2020, Ym_C_A_2020, Zhuang_Zhai_Yamins_2019} have shown that the category-level contrastive learning signal has significant advantages over the instance-level contrastive learning signal in learning distinguishable and invariant features, resulting in better performance in various downstream vision tasks. However, the performance of these methods is heavily dependent on the clustering results. DeepCluster often suffers from the problem of clustering degradation \cite{Walter_Gibson_Sowmya_2020}. That is, due to the poor features used for clustering at the early stage of training or the improper setting of parameters of k-means, all samples may be classified into one cluster and thus mislead feature learning. To avoid clustering degradation, recent studies \cite{Caron_Bojanowski_Joulin_Douze_2018, Ym_C_A_2020} add constraints to the clustering process to make the sample size of each category as balanced as possible. Besides, during training, the above two-step methods need to frequently traverse the entire dataset for offline clustering, which limits the application to large-scale datasets. To solve this issue, Caron et al., \cite{Caron_Misra_Mairal_Goyal_Bojanowski_Joulin_2020} proposed an online clustering-based method, SwAV, that combines the two processes of clustering and cluster assignments prediction into a classification task.
Further studies \cite{Caron_Touvron_Misra_Jegou_Mairal_Bojanowski_Joulin_2021, Li_Hu_Liu_Peng_Zhou_Peng_2021, Li_Zhou_Xiong_Hoi_2021, Wang_Liu_Yu_2021, Zhou_Zhou_Si_Yu_Ng_Yan_2022} found that joint category-level and instance-level contrastive learning (i.e., multi-granular contrastive learning) has higher generalizability than single-granular contrastive learning on diverse downstream tasks. The main reason is that various downstream tasks often require multi-granular features. For example, in the object detection task dataset, DOTA, one needs coarse-grained features to distinguish between bridges and aircraft, and fine-grained features to distinguish sub-categories of aircraft, such as Boeing 737 and Boeing 747. Therefore, multi-granular contrastive learning has become a focus of the studies relating to general-purpose SSFL. Representative approaches of multi-granular contrastive learning \cite{Cao_Li_Feng_Chen_Xia_Zhao_2021, Hu_Li_Zhou_Peng_2021, La_Rosa_Oliveira_Ghamisi_2022, Li_Qin_Ling_Wang_Lin_An_2021, Saha_Shahzad_Mou_Song_Zhu_2022} include PCL \cite{Li_Zhou_Xiong_Hoi_2021} and Mugs \cite{Zhou_Zhou_Si_Yu_Ng_Yan_2022}.

\textbf{Time-series-level contrastive learning signal} aims at learning temporal-invariant features of RSIs \cite{Akiva_Purri_Leotta_2022, Ayush_Uzkent_Meng_Tanmay_Burke_Lobell_Ermon_2021, Chen_Zhang_Hong_Chen_Yang_Li_2022, Chen_Zao_Liu_Chen_Shi_2022, Dong_Ma_Wu_Zhang_Jiao_2020, Leenstra_Marcos_Bovolo_Tuia_2021, Poppelbaum_Chadha_Schwung_2022}. It takes RSIs of different time phases in the same region as positive samples and RSIs of different regions as negative samples to construct the learning signal for learning temporal-invariant features. The motivation of this learning signal is that multi-temporal RSIs have inherent temporal self-similarity. In other words, ground objects with close geospatial distance and temporal phase should be similar. As a representative method, Ayush et al. \cite{Ayush_Uzkent_Meng_Tanmay_Burke_Lobell_Ermon_2021} proposed the geography-aware contrastive learning method to construct positive samples using spatially aligned images over time instead of different augmented views of the same image. They argued that this learning signal makes the learned features more invariant to subtle temporal changes (e.g., due to imaging conditions), resulting in higher distinguishability for spatial variation, thus benefiting target detection and land cover classification tasks. In addition, considering the seasonal visual differences of some ground objects (e.g., vegetation, cropland), Manas et al. \cite{Manas_Lacoste_Giro-i-Nieto_Vazquez_Rodriguez_2021} proposed the seasonal contrast framework (SeCo) to learn the temporal-invariant features. SeCo collects Sentinel-2 images from the four seasons globally and considers the images of the same region in different seasons as positive samples and images from different regions as negative samples. The seasonal-invariant features are learned by the model by distinguishing between positive and negative sample representations. However, as the temporal variation of surface coverage, RSIs of the same area at different time phases may contain completely different land use or land cover types, but they still are treated as positive samples by the above method, which would reduce the distinguishability of the learned features. To construct positive samples that always have the same semantic content while representing various spatio-temporal distributions, Huang et al. simulate RSIs as different spatio-temporal visual styles and used them as positive samples by introducing an optimal transmission mechanism \cite{Huang_Mou_Li_Li_Chen_Li_2022}. In this way, the SSFL model can obtain features with both distinguishability and spatio-temporal invariance and thus better generalize to unseen RSI scene classification datasets.

\begin{figure*}[!t]
  \centering
  \includegraphics*[width=0.75\linewidth]{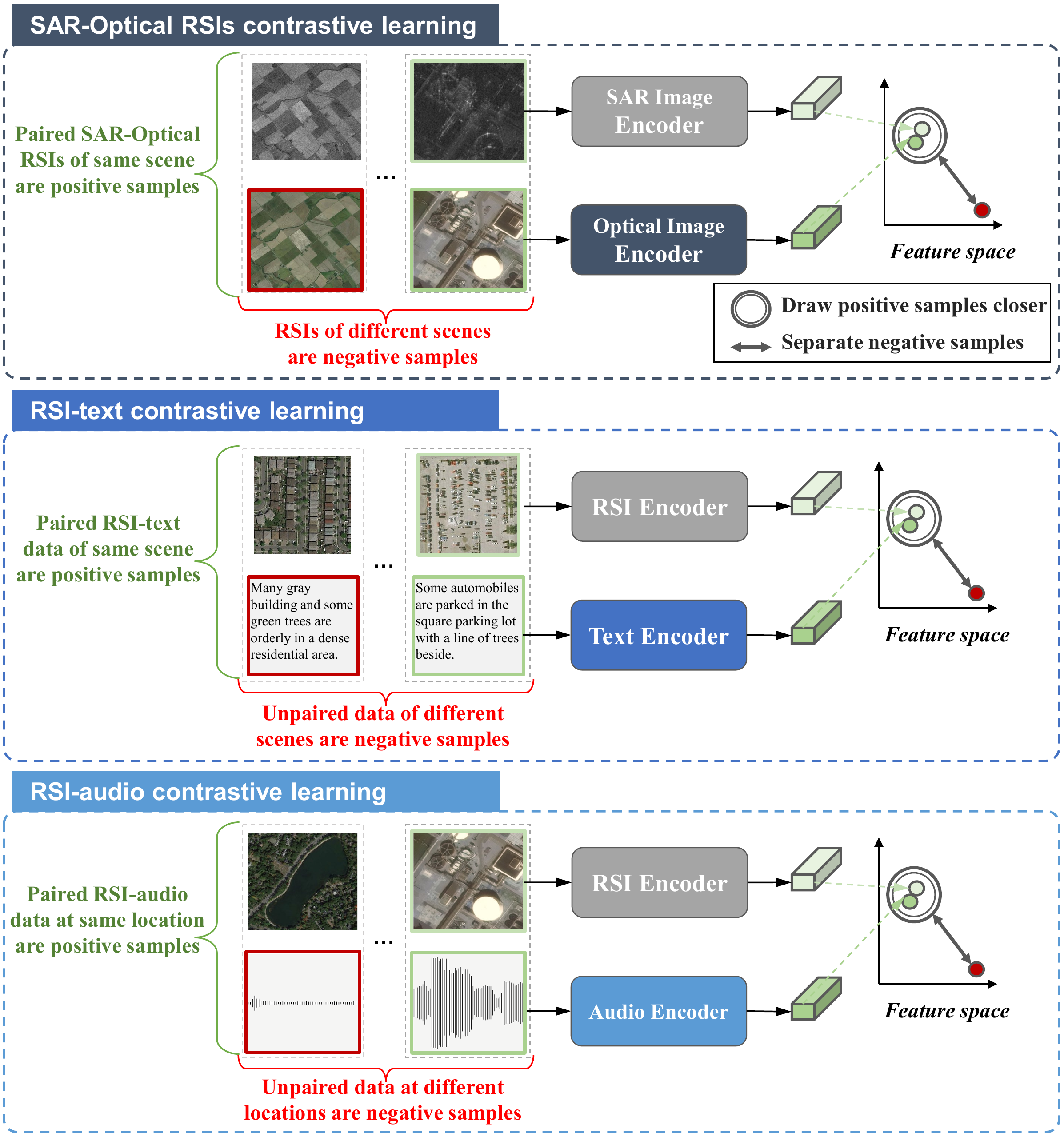}
  \caption{Illustration of three kinds of multi-modal data contrastive learning signals: SAR-optical, RSI-text, and RSI-audio.}
  \label{fig:ssl_contrastive_multi_modal}
\end{figure*}  

\paragraph{\textbf{Multi-modal contrastive learning}}\label{sec:ssl_signal_contrastive_multi_modal}
We can get multi-modal data for the same scene in the same region. These different modal data have great visual differences, but semantically they are the different view representations of the same scene, so they contain embedded invariant features for the same scene. Furthermore, due to the different imaging mechanisms, different modal data of the same scene have complementary features to each other. Huang et al. \cite{Huang_Du_Xue_Chen_Zhao_Huang_2021} theoretically and experimentally demonstrated that the more modal data used, the more likely the deep learning model to gain high-quality complementary features and achieve better performance in classification tasks. Therefore, the multi-modal contrastive learning method, which takes multi-model data of the same scene as positive samples and data of different scenes as negative ones, has attracted great attention \cite{Akbari_Yuan_Qian_Chuang_Chang_Cui_Gong_2021, Arandjelovic_Zisserman_2017, Lan_Liu_Lin_2022, Li_Gao_Niu_Xiao_Liu_Liu_Wu_Wang_2021, Miech_Alayrac_Smaira_Laptev_Sivic_Zisserman_2020, Tian_Krishnan_Isola_2020, Zhou_Brown_Snavely_Lowe_2017}. As shown in Figure \ref{fig:ssl_contrastive_multi_modal}, the method has been applied to learn the cross-modal features of different types of RSIs (e.g., SAR-optical RSIs contrastive learning \cite{Wang_Albrecht_Zhu_2022, Saha_Ebel_Zhu_2022, Montanaro_Valsesia_Fracastoro_Magli_2022, Jain_Schoen-Phelan_Ross_2022, Chen_Bruzzone_2022a, Chen_Bruzzone_2022b, Cha_Seo_Choi_2022, Jain_Schoen-Phelan_Ross_2021, Chen_Bruzzone_2021}) or the combinations of RSIs and other types of data (e.g., RSI-text, and RSI-audio contrastive learning \cite{Heidler_Mou_Hu_Jin_Li_Gan_Wen_Zhu_2021, Mikriukov_Ravanbakhsh_Demir_2022b, Mikriukov_Ravanbakhsh_Demir_2022a}).

\begin{itemize}
  \item \textbf{SAR-optical contrastive learning}: Optical RSIs have rich texture details, but the image quality is susceptible to cloud and rain interference. In contrast, SAR satellites provide consistent quality observations in a variety of weather and lighting conditions. Therefore, combining SAR and optical RSIs is an important issue for multi-modal RSI contrastive learning. Chen and Bruzzone \cite{Chen_Bruzzone_2022a} took pairs of heterogeneous images (SAR-optical) from the same location as positive samples, and images from different locations as negative samples. Considering the different data structures of SAR and optical images, the method first constructs two independent encoders to extract features from the two types of images separately and then uses the multi-modal contrastive learning signal to train the encoders for SSFL. The performance of the learned features was assessed using the heterogenous RSI change detection tasks for two SAR-optical image datasets (including the Sentinel-1/2 dataset and the Sentinel-1/Landsat-8 dataset). Moreover, Chen and Bruzzone \cite{Chen_Bruzzone_2022b} compared the performance of multi-modal (SAR-optical) and single-modal (SAR or optical) SSFL models in three land-cover classification tasks and found that the multi-modal contrastive learning approach outperforms single-modal contrastive learning approaches.
  \item \textbf{RSI-text contrastive learning}: The research of contrastive learning based on image-text data develops rapidly in the field of computer vision \cite{Jia_Yang_Xia_Chen_Parekh_Pham_Le_Sung_Li_Duerig_2021, Li_Yatskar_Yin_Hsieh_Chang_2019, Lu_Batra_Parikh_Lee_2019, Zhang_Jiang_Miura_Manning_Langlotz_2020}, because a massive amount of paired natural images and corresponding caption text can be automatically crawled from the Internet at a low cost. As a millstone work, Contrastive Language Image Pre-training (CLIP) \cite{Radford_Kim_Hallacy_Ramesh_Goh_Agarwal_Sastry_Askell_Mishkin_Clark_2021} constructed a large-scale Internet crawler dataset containing 400 million image-text sample pairs for multi-modal SSFL. CLIP constructs this learning signal by taking paired images and text as positive samples, and unpaired images and text as negative samples. During the pre-training, CLIP feeds image and text samples into the image encoder and text encoder for independent encoding, which generates uniformly structured feature vectors. Then, the contrastive learning signal is used to drive the encoders to learn cross-modal features. By the above pre-training, CLIP can directly generate textual descriptions that accurately describe the semantics of the input images. Based on predefined mapping rules, these descriptions can be associated with the categories of the downstream dataset for classification or recognition. Without using any labeled sample for finetuning, the multi-modal CLIP matches the accuracy of the original ResNet50 supervised trained with 1.28 million labeled images on the ImageNet classification task. However, there are few relevant studies in the field of remote sensing since textual descriptions of RSIs are difficult to obtain. So far, the only work is done by Mikriukov et al \cite{Mikriukov_Ravanbakhsh_Demir_2022a}, which used two publicly available RSI captions datasets\footnote{https://github.com/201528014227051/RSICD\_optimal}, UC Merced Land Use and RSICD. The results of both two datasets showed that the model trained by the learning signal can accurately retrieve the corresponding RSIs based on the input text caption. However, the text descriptions of RSIs in these two datasets are obtained by manual annotation, so this paradigm is hard to be expended. Therefore, it is worth studying how to obtain geo-tagged textual information of RSIs from crowed-sourced data such as OpenStreetMap and Twitter at a low cost.
  \item \textbf{RSI-audio contrastive learning}: Multi-modal contrastive learning based on image-audio has also attracted the attention of scholars \cite{Arandjelovic_Zisserman_2017, Chen_Xie_Vedaldi_Zisserman_2020, Lan_Liu_Lin_2022, Owens_Wu_McDermott_Freeman_Torralba_2016}. In an open-source project called Radio Aporee:::Maps\footnote{https://archive.org/details/radio-aporee-maps}, Heidler et al. \cite{Heidler_Mou_Hu_Jin_Li_Gan_Wen_Zhu_2021} constructed a multi-modal dataset called SoundingEarth containing 50,545 RSI-audio sample pairs. Specifically, they collected live sounds recorded by volunteers around the world in residential areas, parks, lakes, and wilderness scenes with corresponding high-resolution RSIs from the Google Earth platform. On the basis of SoundingEarth, they used an approach similar to RSI-text multi-modal SSFL for RSI-audio feature learning. Experiments demonstrate that introducing audio data into feature learning is also useful for downstream vision tasks, such as remote sensing scene classification and land cover classification.
\end{itemize}

\subsection{Evaluation methods for self-supervised feature learning}\label{sec:ssl_evaluation}
The representation capability of the features learned from massive remote sensing data in a self-supervised manner should be assessed accurately and objectively. Specifically, do these learned features have strong distinguishability in remote sensing interpretation tasks? Are they invariant for remote sensing images of different regions, time phases, and resolutions? The performance of SSFL can be evaluated qualitatively and quantitatively.

\subsubsection{Qualitative evaluation methods}
The qualitative evaluation method visualizes the learned feature to evaluate the quality. There are three commonly used methods, including Kernel Visualization \cite{Jing_Yang_Liu_Tian_2019}, Feature Map Visualization \cite{Selvaraju_Cogswell_Das_Vedantam_Parikh_Batra_2020, Wang_Wang_Du_Yang_Zhang_Ding_Mardziel_Hu_2020, Zeiler_Fergus_2014}, and T-SNE \cite{Maaten_Hinton_2008} unsupervised clustering visualization. Feature Map Visualization got a wide application, which leverages technologies such as deconvolution and class activation visualization to visualize the activation feature map of the input image obtained by the SSFL model. On this basis, we can observe which regions the model pays attention to when understanding an input image, and are these regions consistent with those attracting human attentions?

The literature \cite{Caron_Touvron_Misra_Jegou_Mairal_Bojanowski_Joulin_2021}, using the Feature Map Visualization method, assessed the quality of the features obtained from natural image datasets by self-supervised learning based on the vision transformation (VIT) framework and compares those features with those obtained by supervised methods. The experimental results showed that the features learned by the VIT framework in a self-supervised manner without using any category labels can focus on the most relevant regions of the image's semantic meaning, and their interpretability is even better than the features got by the supervised feature learning method. In addition, those features form more separated clusters, indicating that those features have strong category distinguishability. The literature \cite{Zhao_Luo_Li_Chen_Piao_2020} used a similar visual analysis method to assess the quality of the features obtained by the jointly supervised feature learning method and the self-supervised feature learning method from a remote sensing scene classification dataset. The results showed that combining two feature learning paradigms can better capture the details that represent the semantic features in the images with complex backgrounds, and achieve higher accuracy in classifying some difficult objects than the baseline model does.

\subsubsection{Quantitative evaluation methods}
At present, the self-supervised learned features are usually assessed quantitatively by downstream tasks. Specifically, the features obtained by SSFL are used as pre-trained model parameters and are transferred to downstream tasks (e.g., remote sensing scene classification, semantic segmentation, and target detection). Then their performance in downstream tasks is evaluated and the results are taken as the assessment results of the features. The commonly used transfer methods are linear probes and fine-tuning.

The process of the linear probe method is: 1) the network parameters obtained by self-supervised learning are fixed and a linear classifier is added to the last layer of the network. 2) The linear classifier is trained using downstream labeled data to evaluate the performance of self-supervised learning features. However, due to the simple classification structure, this method can only evaluate the performance of self-supervised learning features in image-level classification tasks (e.g., remote sensing scene classification). The literature \cite{Tao_Qi_Lu_Wang_Li_2022} used the linear probe approach to compare the representation ability of the models pre-trained by SSFL methods for three popular RSIs scene classification datasets. The experiments showed that the features learned by the instance-level contrastive SSFL method have better performance in scene classification tasks than the methods based on image inpainting or predicting relative position.

\begin{figure}[!t]
  \centering
  \includegraphics*[width=0.98\linewidth]{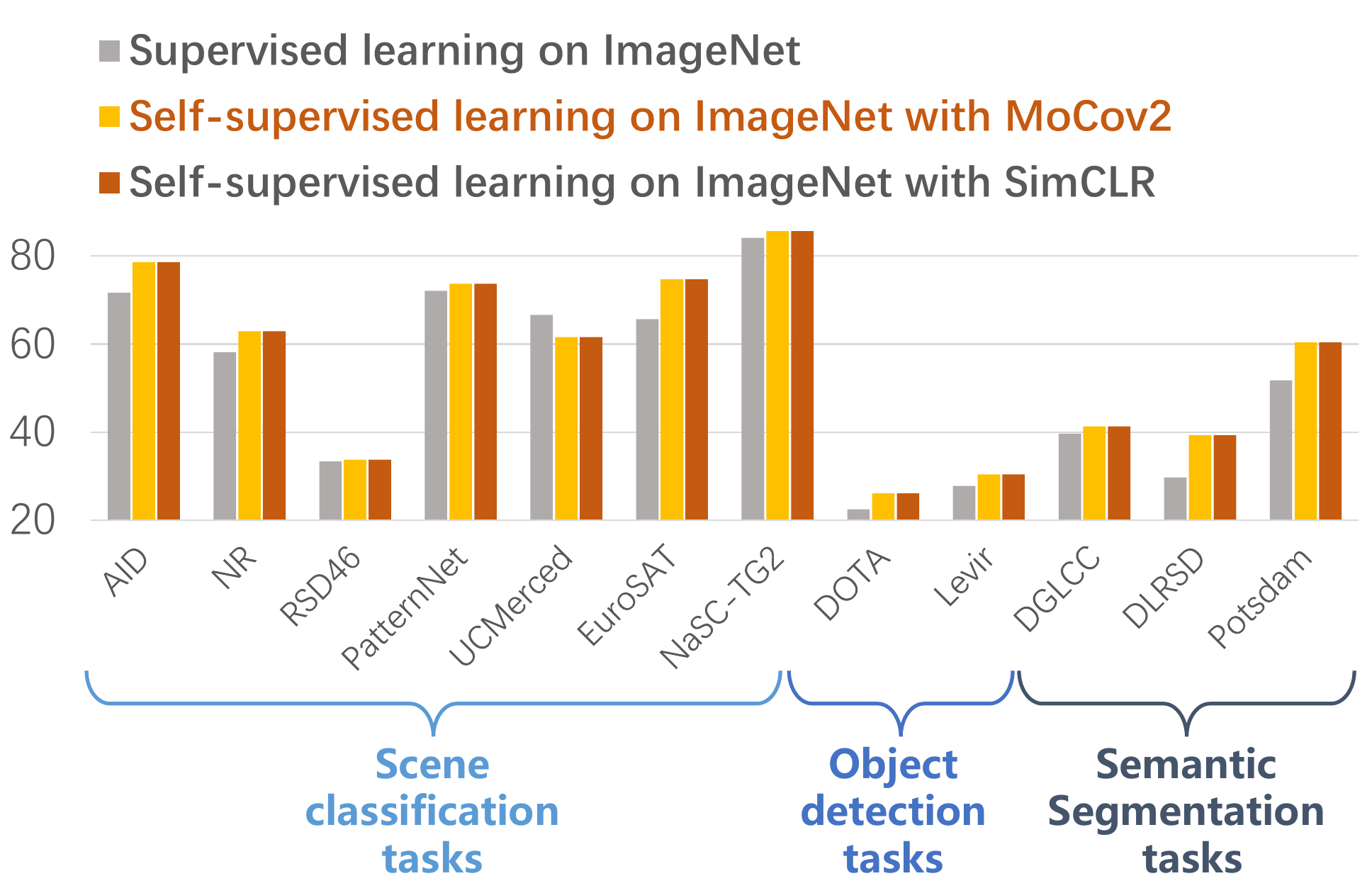}
  \caption{Quantitative evaluation of self-supervised and supervised learned features on multiple downstream tasks by fine-tuning approach. To evaluate the performance of the features learned by different methods, we adopt three types of RSI understanding tasks, including scene classification, object detection, and semantic segmentation. Scene classification tasks used AID, NR, RSD46, PatternNet, UC Merced, EuroSAT, and NaSC-TG2. Object detection tasks used DOTA and LEVIR. Semantic segmentation tasks include DGLCC, DLRSD, and Potsdam.}
  \label{fig:tov_results}
\end{figure}  

The fine-tuning approach uses the model parameters pre-trained by SSFL methods as an initialization of the task-specific model’s backbone, then transfers the learned features to various RSIU tasks by adding task-specific adapters following the backbone. Since the features learned by the SSFL method can be adapted to different downstream tasks by simply modifying the task-specific adapters, this feature transfer approach enables a more comprehensive feature representation capability assessment. Tao et al. used the fine-tuning approach to test the representation capability of the features learned by the SSFL method on ImageNet \cite{Tao_Qi_Zhang_Zhu_Lu_Li_2022}, which used 12 publicly available datasets for the downstream tasks of remote sensing scene classification, object detection, and semantic segmentation. The results (Figure \ref{fig:tov_results}) show that in all three types of tasks, the features obtained by the SSFL method perform better than those obtained by the supervised learning method. However, this comparison approach cannot intuitively accomplish a comprehensive comparison between different self-supervised feature methods. For example, SimCLR performs better than MoCov2 for the scene classification task but worse for the target recognition task. Theoretically, self-supervised learning is a task-independent approach, and the generality of the learned features is also very important. Thus, future research could be focused on evaluating the combined performance of the features on multiple tasks.

The literature \cite{Wang_Braham_Albrecht_Xiong_Liu_Zhu_2022} assessed the transferability of self-supervised learned features by the EuroSAT \cite{Helber_Bischke_Dengel_Borth_2019} and BigEarthNet \cite{Sumbul_Charfuelan_Demir_Markl_2019} datasets using the linear probe and fine-tuning. They performed SSFL using the EuroSAT dataset after removing the original annotation and transferred the learned features to the downstream BigEarthNet scene classification dataset. The results show that regardless of the feature transfer method, the self-supervised learning features have good transferability. Notably, the fine-tuned transfer method using only 10\% of the downstream task labels can get a result similar to that using all downstream task labels.

\section{The key factors influencing self-supervised feature learning}\label{sec:exp}
In the past two years, SSFL methods have developed rapidly in the field of remote sensing, but there lacks a comprehensive study on the key factors that would influence SSFL. Therefore, we analyze how the SSFL signal and properties of pre-training data affect the performance of the learned features in downstream tasks in Section \ref{sec:exp_ssl_signal} and \ref{sec:exp_ssl_factors}, respectively.

\subsection{Self-supervised feature learning signals}\label{sec:exp_ssl_signal}

For the downstream tasks with very limited labeled data, optimizing the model based on a good starting point can reduce the risk of overfitting. Therefore, the choice of SSFL signal is crucial, as it determines what features can be learned by pre-training and whether the features are relevant to downstream tasks. This experiment analyzes how the SSFL signal affects the performance of the learned features in downstream tasks.

\begin{table*}[!t]
  \centering
  \renewcommand{\arraystretch}{1.1}
  \caption{Information of the training and testing sets of the six datasets for the downstream tasks in the experiments.}
    \begin{tabular}{ccccccc}
    \toprule
    Dataset & RSD46-WHU & EuroSAT & DOTA & LEVIR & Potsdam & DGLCC \\
    \midrule
    Classes & 46 & 10 & 15 & 3 & 6 & 7 \\
    Training set size & 800 & 800 & 400 & 400 & 200 & 200 \\
    Testing set size & 10000 & 10000 & 5000 & 5000 & 2500 & 2500 \\
    \bottomrule
    \end{tabular}
  \label{tab:downstream_datasets}
\end{table*} 

\subsubsection{Experiment Setup}\label{sec:exp_ssl_signal_info}
Corresponding to the generative, predictive, and contrastive learning signals, we choose the following three SSFL methods for comparison:

\begin{enumerate}[i.]
  \item	Inpainting \cite{Pathak_Krahenbuhl_Donahue_Darrell_Efros_2016}. The model learns features by recovering the manually masked parts.
  \item Image rotation prediction \cite{Gidaris_Singh_Komodakis_2018}. The model learns features by predicting the rotation angle of the input images.
  \item Instance-level contrastive learning \cite{Chen_Kornblith_Norouzi_Hinton_2020}. First, the argument views of the same sample are regarded as positive instances and that of different samples in a training batch are regarded as negative instances. Then, the model learns features by enhancing the similarity between positive instances and the difference between negative instances.
\end{enumerate}

\paragraph{Pre-training datasets}
We pre-train the above three SSFL methods using two large-scale RSI datasets, Million-AID \cite{Long_Xia_Li_Yang_Yang_Zhu_Zhang_Li_2021, Long_Xia_Zhang_Cheng_Li_2022} and TOV-RS \cite{Tao_Qi_Zhang_Zhu_Lu_Li_2022}, separately. As described in Section \ref{sec:ssl_dataset}, Million-AID is a manually collected and labeled dataset, containing 1,000,848 RSI samples of 51 categories with good diversity. We use these samples without labels for pre-training. TOV-RS dataset is an automatically collected unlabeled dataset containing 3 million samples.

\paragraph{Downstream datasets}
To evaluate the performance of the features learned by these SSFL methods, we adopt three types of RSI understanding tasks, which are scene classification, object detection, and semantic segmentation, using different datasets.

\begin{enumerate}[i.]
\item Dataset for scene classification tasks:
\begin{itemize}
  \item RSD46-WHU \cite{Long_Gong_Xiao_Liu_2017, Xiao_Long_Li_Wei_Tang_Liu_2017}, a large-scale scene classification dataset containing 117,000 samples of 46 categories. The images are collected from Google Earth and Tianditu.
  \item	EuroSAT \cite{Helber_Bischke_Dengel_Borth_2019}, a scene classification dataset containing 27,000 Sentinel-2 images of 10 categories.
\end{itemize}

\item Dataset for object detection tasks:
\begin{itemize}
  \item DOTA v1.0 \cite{Ding_Xue_Xia_Bai_Yang_Yang_Belongie_Luo_Datcu_Pelillo_2021, Xia_Bai_Ding_Zhu_Belongie_Luo_Datcu_Pelillo_Zhang_2018}, a large-scale object detection dataset containing 188,282 objects of 15 categories. It consists of RGB images and grayscale images collected from Google Earth, CycloMedia, GF-2, and JL-1 satellite.
  \item LEVIR \cite{Zou_Shi_2018}, is an object detection dataset consisting of over 22,000 Google Earth images with a resolution of 1.0 - 0.2 m/pixels. It has three categories: airplane, ship, and oil tank.
\end{itemize}

\item Dataset for semantic segmentation tasks:
\begin{itemize}
  \item Potsdam \cite{Rottensteiner_Sohn_Jung_Gerke_Baillard_Benitez_Breitkopf_2012}, is one of the most popular semantic segmentation datasets in the RS field. It contains 38 UAV images with a resolution of 0.05 m/pixels and a size of 6000x6000. This dataset is annotated in 6 classes.
  \item DGLCC \cite{Demir_Koperski_Lindenbaum_Pang_Huang_Basu_Hughes_Tuia_Raskar_2018}, a land cover classification dataset containing 1,146 images annotated in 7 classes. These images are collected by the DeepGlobe satellites. The images have a size of 2448×2448 and a resolution of 0.5 m/pixels. These images are mainly located in Thailand, Indonesia, and India, covering a total area of 1716.9 $\text{km}^2$.
\end{itemize}
\end{enumerate}

Details of the training and testing set of the above datasets are shown in Table \ref{tab:downstream_datasets}.

\begin{figure}[!t]
  \centering
  \includegraphics*[width=0.99\linewidth]{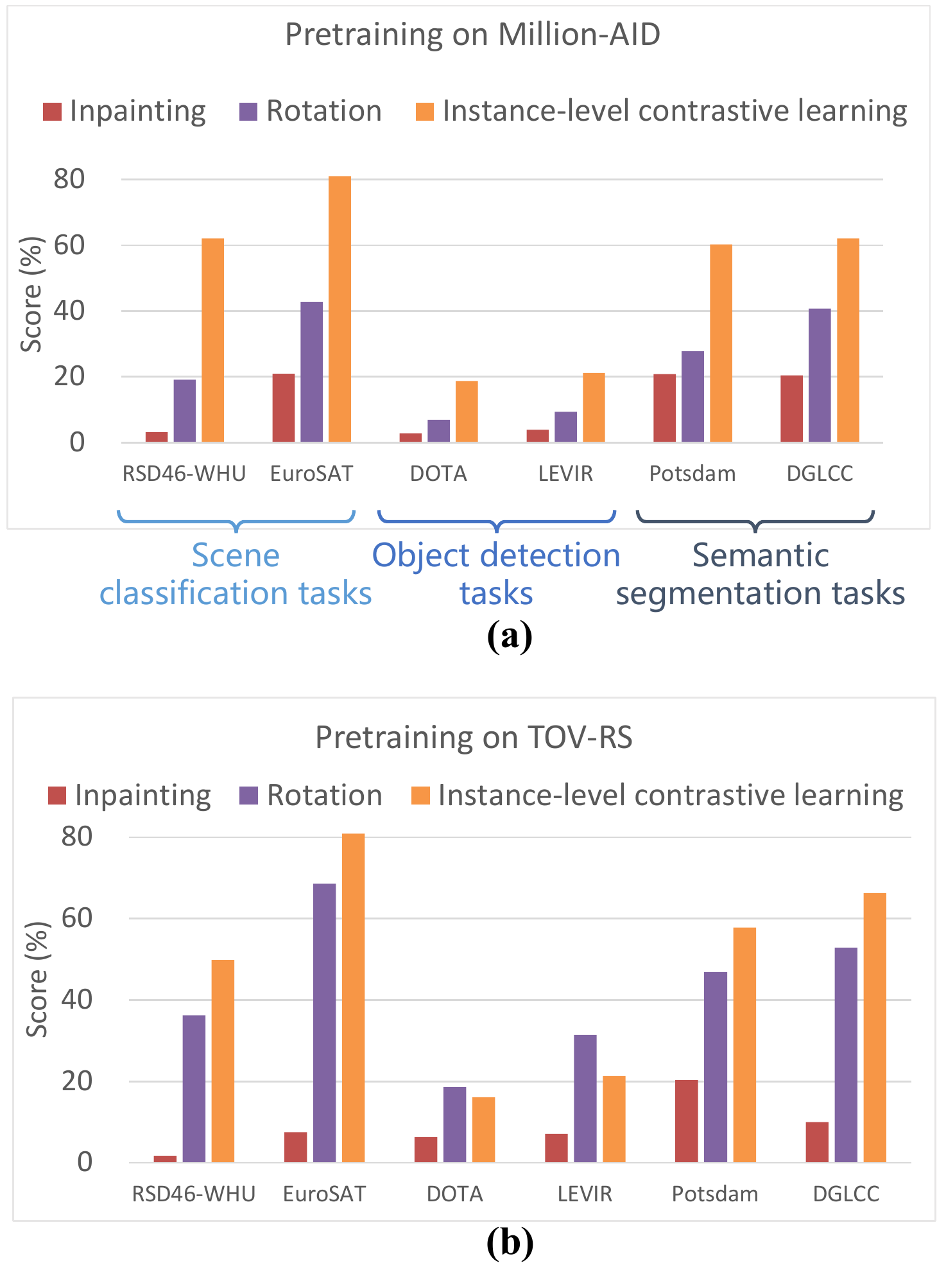}
  \caption{Performance of the ResNet50 model pre-trained by three self-supervised feature learning signals using (a) Million-AID and (b) TOV-RS. The performance using RSD46-WHU, EuroSAT, Potsdam, and DGLCC is assessed by the Kappa value, and that using DOTA and LEVIR is evaluated by the mAP value.}
  \label{fig:exp_signal}
\end{figure}  

\paragraph{Implementation details}
 We pre-train the ResNet50 \cite{He_Zhang_Ren_Sun_2016} model for 400 epochs using the three SSFL methods on two pre-training datasets separately and then evaluate the six pre-trained models by different downstream tasks. We follow the default hyperparameter settings of the official repository of the three SSFL methods and use 2 NVIDIA A100 GPUs for the experiments, on which the batch size is 256.

For evaluation, the pre-trained models are applied to six downstream task datasets by fine-tuning them on the training set of each dataset (Table \ref{tab:downstream_datasets}). The performance in scene classification and semantic segmentation is assessed by Cohen’s kappa coefficient (kappa), and that in object detection is assessed by the mean average precision (mAP).

\subsubsection{Experiment results}\label{sec:exp_ssl_signal_result}
Below are our findings:
\begin{itemize}
  \item \textbf{Contrastive learning is an optimal choice because the learned features are superior in most downstream tasks}. As shown in Figure \ref{fig:exp_signal}(a), when pre-trained by Million AID, the performance scores of instance-level contrastive learning are much higher than other types of SSFL methods in all six downstream tasks. When pre-trained by TOV-RS (Figure \ref{fig:exp_signal}(b)), instance-level contrastive learning also achieves the highest scores in scene classification and semantic segmentation tasks. Therefore, the contrastive learning signal is a promising solution for multiple RSI understanding tasks or uncertain downstream tasks.
  \item \textbf{The choice of SSFL signal should consider the downstream tasks}. For the two object detection tasks, as shown in Figure \ref{fig:exp_signal}(b), the image rotation prediction method achieves higher scores than the instance-level contrastive learning method. This suggests that the rotation-equivariant features learned by image rotation prediction signals are important for object detection tasks. In contrast, the instance-level contrastive learning aiming at distinguishing RSI scenes may not learn the object-level distinguishable feature. This confirms the gap between instance-level contrastive learning and object detection tasks mentioned in Section \ref{sec:ssl_signal_contrastive_single_modal}.
\end{itemize}

\subsection{Effects of the pre-training datasets on the performance of the self-supervised learned features in downstream tasks}\label{sec:exp_ssl_factors}
Spatial resolution and data volume are the two basic and important properties of a dataset. The former determines the richness of the spatial information that the model learns from RSIs. The latter determines the diversity of the learned features. Thus, the following experiments analyze how these two properties affect the performance of the learned features in downstream tasks.

\subsubsection{Study on spatial resolution}\label{sec:exp_ssl_factors_resolution}
\paragraph{Experiment Setup}
We use instance-level contrastive learning for feature learning, as its superiority has been demonstrated in the above experiments.
 To investigate the effects of spatial resolution, we use one medium-low resolution dataset (SeCo) and two high-resolution datasets (TOV-RS and Million-AID) for pre-training. SeCo has a spatial resolution of up to 10 m/pixel, and the TOV-RS and Million-AID have resolutions of up to 1 m and 0.5 m, respectively.
To evaluate the performance of the features learned by the model pre-trained by different datasets, we use the six downstream datasets described in Section \ref{sec:exp_ssl_signal_info} for downstream tasks.

\begin{figure}[!t]
  \centering
  \includegraphics*[width=0.98\linewidth]{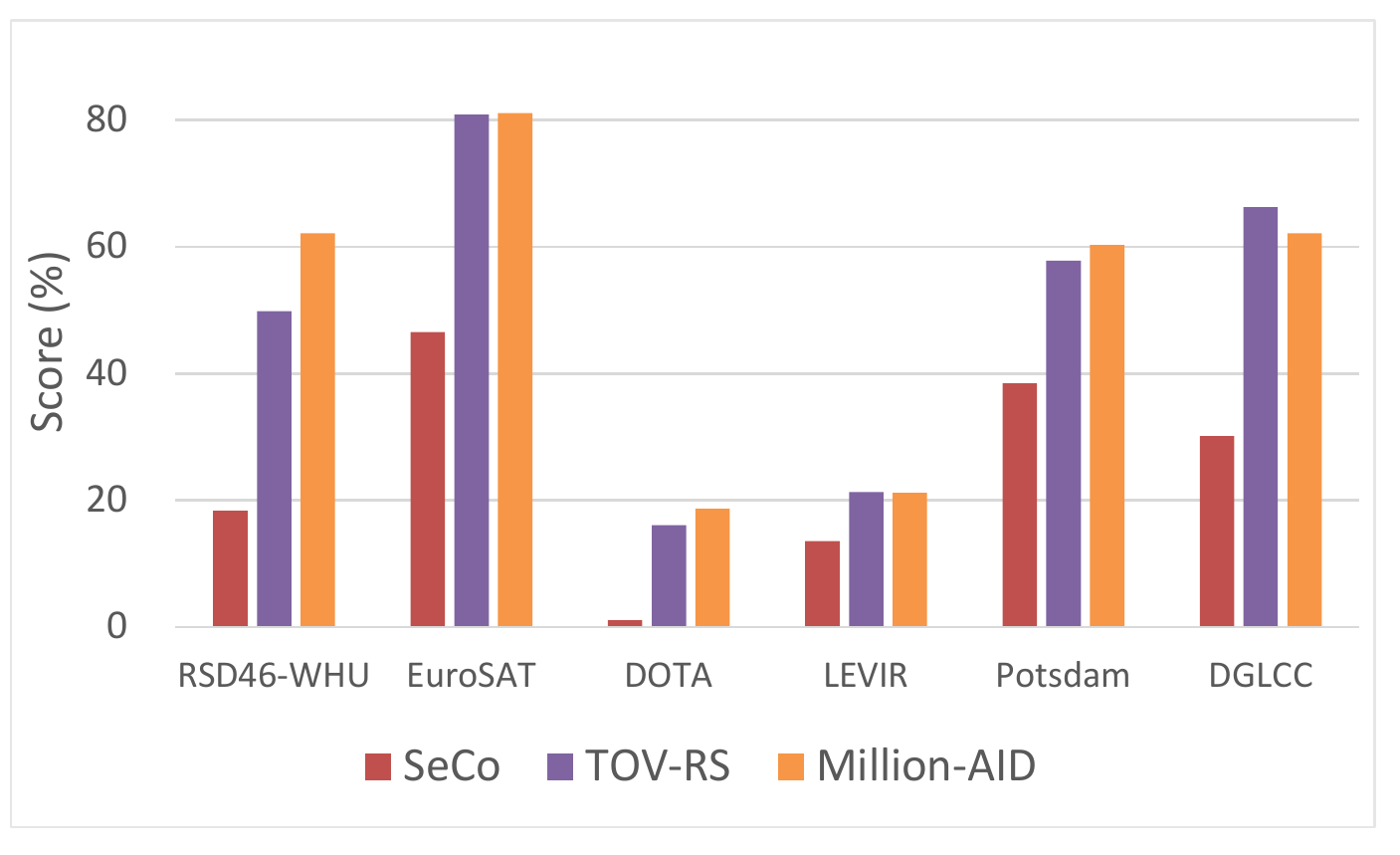}
  \caption{Results for the data of different spatial resolutions. RSD46-WHU, EuroSAT, Potsdam, and DGLCC are assessed by Kappa. DOTA and LEVIR are assessed by mAP.}
  \label{fig:exp_resolution}
\end{figure}  

\paragraph{Experiment results}
Results for the data of different spatial resolutions are shown in Figure \ref{fig:exp_resolution}. Below are our findings:

\begin{itemize}
\item \textbf{Spatial resolution of the pretraining dataset is critical for SSFL}. For all three kinds of downstream tasks, the performance of the model pre-trained using the two high-resolution datasets (TOV-RS and Million-AID) is significantly higher than that of the model trained by the medium-low resolution dataset (SeCo). The reason may be that the high-resolution RSIs dataset can additionally provide more textural and geometric details of the ground objects than the low-resolution dataset, thus the learned features are more distinguishable for various objects and scenes.
\item \textbf{The impact of the pre-training data resolution on self-supervised feature learning may be greater than that of the domain gap between the pre-training and downstream datasets}. For example, the EuroSAT dataset and the SeCo are constructed by the RSIs from Sentinel-2 data, which have a significant domain difference from TOV-RS and Million-AID which are constructed based on Google Earth data. However, as shown in Figure \ref{fig:exp_resolution}, models pre-trained on TOV-RS and Million-AID achieve much higher performance scores than the model pre-trained on SeCo. For EuroSAT, the kappa of the TOV-RS pre-training and Million-AID pre-training models are 0.8089 and 0.8111, respectively, while the kappa of the SeCo pre-training model was only 0.4658. This suggests that it is unnecessary to sacrifice the resolution of the pre-training data for reducing the domain gap between the pre-training data and the downstream task data.
\end{itemize}

In summary, we recommend using RSI datasets with as high resolution as possible for SSFL pretraining.

\begin{figure}[!t]
  \centering
  \includegraphics*[width=\linewidth]{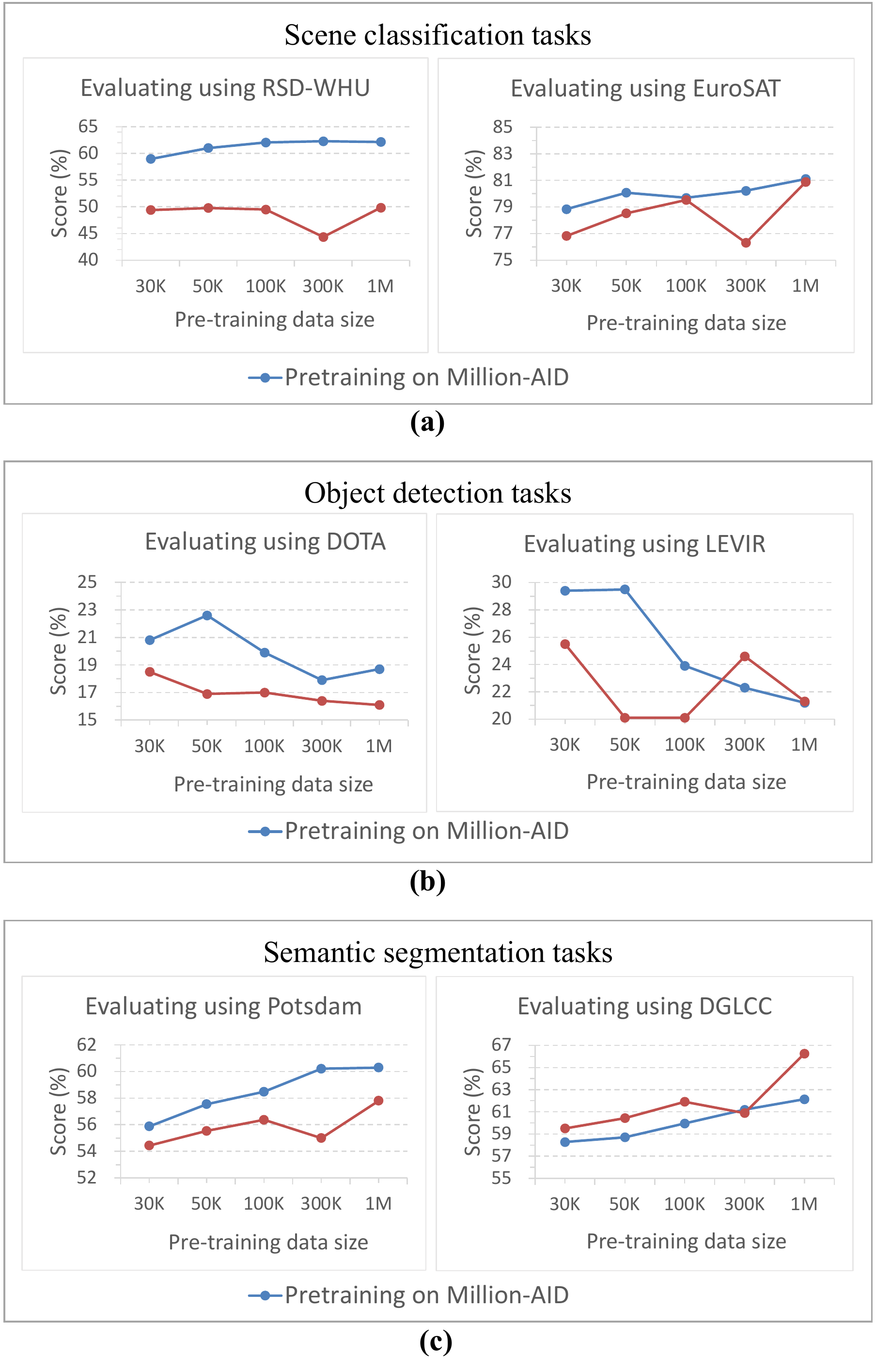}
  \caption{To evaluate the performance of the features learned from pre-training data of different sizes, we use three types of downstream tasks, including (a) scene classification tasks, (b) object detection tasks, and (c) semantic segmentation tasks.}
  \label{fig:exp_datasize}
\end{figure}  

\subsubsection{Study on data size}\label{sec:exp_ssl_factors_datasize}
\paragraph{Experiment Setup}
To analyze how pre-training data size affects the performance of learned features in downstream tasks, we randomly sample subsets of 30K, 50K, 100K, 300K, and 1M images from each of Million-AID and TOV-RS. The other settings are the same as those in Section \ref{sec:exp_ssl_factors_resolution}.

\paragraph{Experiment results}
Results for different pre-training data sizes are shown in Figure \ref{fig:exp_datasize}. Below are our findings:
\begin{itemize}
\item Enlarging the pre-training data size improves the learned features’ performance in most downstream tasks. In Figure \ref{fig:exp_datasize} (a) and (c), as the pre-training data size grows from 30K to 1M, the performance of the learned features in both scene classification and semantic segmentation tasks improves accordingly. The model pre-trained on 1M Million-AID samples outperforms the model pre-trained on 30K Million-AID samples, with kappa increases of 5\% and 8\% for RSD46-WHU and Potsdam, respectively. This phenomenon is not surprising, as deep learning methods are data-hungry \cite{Sun_Shrivastava_Singh_Gupta_2017}. However, for two object detection tasks, as shown in Figure \ref{fig:exp_datasize} (b), the performance of the model shows irregular fluctuations as the pre-training data size increases. This confirms the findings in Section \ref{sec:exp_ssl_signal_result} that the features learned by instance-level contrastive learning that aims at discriminating RSI scenes are unsuitable for object detection tasks.
\item Little benefit over 300,000 pre-trained samples. The performance of the learned features in these tasks seems to saturate before 1M. For example, as shown in Figure \ref{fig:exp_datasize}(a) and (c), when pre-trained on Million-AID, the performance difference between 300,000 and 1,000,000 samples is smaller than 1\% in both scene classification and object detection tasks. That is, 70\% of the one-million image dataset contributes less than 1\% of the accuracy improvement, which can be sacrificed for a remarkable efficiency improvement.
\end{itemize}

\section{Future work}\label{sec:future_work}
\subsection{Theoretical foundation of the internal relationship between pretext tasks and the performance of SSFL}
SSFL methods learn features by the learning signals constructed by multiple pretext tasks, which is different from the supervised feature learning methods using only the class priors from data labels as the feature learning signal. Current research \cite{Ericsson_Gouk_Hospedales_2021, Goyal_Mahajan_Gupta_Misra_2019, Kolesnikov_Zhai_Beyer_2019, Li_Chen_Shi_2021, Newell_Deng_2020, Tao_Qi_Lu_Wang_Li_2022, Zhai_Puigcerver_Kolesnikov_Ruyssen_Riquelme_Lucic_Djolonga_Pinto_Neumann_Dosovitskiy_2020} have shown that the choice of learning signals is crucial for the performance of the learned features in downstream tasks, and the learning signal constructed by a poor pretext task may hardly learn the features that can promote the downstream tasks. However, the intrinsic relationship between the learning signal and the feature representation capability is still unclear. What kind of features can be learned from remote sensing data by existing self-supervised learning methods? Are the features learned by different SSFL signals different? If so, is it possible to combine different SSFL signals to achieve complementary feature learning? In addition, does design an SSFL signal relevant to the remote sensing interpretation task can improve the feature representation capability? To answer the above questions, it is necessary to carry out related theoretical studies to improve the understanding of the current SSFL paradigms and their intrinsic mechanisms, as well as to provide theoretical guidance for designing better self-supervised pretext tasks.

\subsection{Transfer self-supervised learned features to downstream tasks}
SSFL signals are designed with various motivations, so the correlation between the learned features and the downstream target tasks may be different. If the feature is regarded as a kind of knowledge, the feature that is strongly associated with the target task may be a kind of “special knowledge”, and the feature that is weakly associated with the target task may be a kind of “general knowledge”. It means that they have different effects on downstream tasks. Thus, using a unified feature transfer strategy, either linear probe or fine-tuning, for the features learned by different SSFL methods may result in the ineffective transfer or even negative transfer, and consequently weaken the generalization performance of the model. Therefore, further research should be done on self-supervised learning feature transfer methods. For example, 1) propose criteria to measure the correlation between self-supervised learned features and downstream tasks, and then design feature transfer methods according to their relationship. 2) develop end-to-end transfer methods from self-supervised feature learning to supervised downstream tasks, allowing the network to learn features adaptively to fit the downstream tasks.

\subsection{Continual self-supervised feature learning model from multimodal remote sensing data}
With the rapid development of the Global Earth Observation System of Systems (GEOSS), a huge amount of remote sensing data become available. To process the constantly growing unlabeled remote sensing data, it is necessary to achieve self-growth feature representation capability through continual self-supervised learning from streaming remote sensing data. In addition, the remote sensing data may be acquired by multiple sensors and in different modalities (e.g., hyperspectral, multispectral, SAR, satellite video), so constructing models for each modality will increase the training cost and may not be able to learn complementary features. Although a lot of methods related to multi-modal SSFL have been developed in the field of computer vision \cite{Arandjelovic_Zisserman_2017, Radford_Kim_Hallacy_Ramesh_Goh_Agarwal_Sastry_Askell_Mishkin_Clark_2021, Tian_Krishnan_Isola_2020}, most of these methods are based on the assumption that the data from different modalities are precisely aligned. So, they are seldomly used in the field of remote sensing. The precise spatial-temporal alignment between different modalities of data is a technical problem that has not been solved \cite{Xu_Ma_Yuan_Le_Liu_2022}. Therefore, for the unaligned data without annotation, how to construct continual feature learning paradigms for multi-modal data and learn the intrinsic relationship between different modalities are worth researching in the future.

\subsection{The benchmark for remote sensing SSFL evaluation}
As described in Section \ref{sec:ssl_evaluation}, the quality of the self-supervised learned remote sensing features is generally assessed by downstream RSI understanding tasks. However, there is still a lack of a standardized benchmark for SSFL evaluation in the field of RS. In addition, it is important to measure the generality of self-supervised learned features on different downstream tasks. However, the evaluation indicators for different tasks are different, for example, scene classification uses overall accuracy, and object detection generally uses mean average precision. So it is difficult to directly aggregate them statistically (such as averaging) to form a comprehensive evaluation indicator. Therefore, the construction of a general evaluation benchmark and comprehensive evaluation indicators for remote sensing SSFL is important.

\section{Conclusions}\label{sec:conclusion}
In this survey, we provide a unified feature learning framework to link different remote sensing feature learning paradigms. Based on this framework, we analyzed self-supervised remote sensing feature learning from three perspectives: training data, learning signal, and evaluation metrics. Under such a taxonomy, we provided a systematic review of existing research across these categories. Moreover, we performed a comprehensive comparative study to analyze the impacts of training data selection and the choice of SSFL signals on self-supervised remote sensing feature learning. This study can help to foster the development of SSFL in the remote sensing community. Finally, we discuss some problems that should be solved in future research to improve the development of self-supervised remote sensing feature learning.

\section*{Acknowledgment}\label{sec:acknowlegements}
The work presented in this paper was supported by the National Key Research and Development Program (grant number 2018YFB0504501); The National Natural Science Foundation of China (No. 42171376, 41771458, 41871364); The Natural Science Foundation of Hunan (2021JJ30815); The Young Elite Scientists Sponsorship Program by Hunan province of China (No. 2018RS3012); Hunan Science and Technology Department Innovation Platform Open Fund Project (18K005); and the High Performance Computing Center of Central South University.

\bibliographystyle{IEEEtran}
\bibliography{IEEEabrv,Reference_SSL}



\ifCLASSOPTIONcaptionsoff
\newpage
\fi



\end{document}